\pdfoutput=1
\documentclass[a4paper,11pt]{article}
\usepackage{eacl2017}
\usepackage{times}
\usepackage{url}
\usepackage{latexsym}
\usepackage{graphicx}
\usepackage{capt-of}
\usepackage{floatrow}
\usepackage{multirow}
\usepackage[compact]{titlesec}
\usepackage{booktabs}
\usepackage{adjustbox}
\usepackage{array}
\usepackage{multirow}

\newcolumntype{R}[2]{%
    >{\adjustbox{angle=#1,lap=\width-(#2)}\bgroup}%
    l%
    <{\egroup}%
}
\newcommand*\rot{\multicolumn{1}{R{45}{.01em}}}
\usepackage{todonotes}
\newcommand{\Note}[2]{}

\newcommand{\NoteAB}[1]{\Note{blue!40}{#1 --Adrian}}
\newcommand{\NoteMM}[1]{\Note{red!40}{#1 --Meg}}
\newcommand{\NoteDH}[1]{\Note{green!40}{#1 --Dirk}}

\eaclfinalcopy 
\def\eaclpaperid{180} 

\title{Multi-Task Learning for Mental Health\\using Social Media Text}

\author{Adrian Benton\\Johns Hopkins University\\\texttt{\small adrian@cs.jhu.edu} \And Margaret Mitchell\\Microsoft Research\thanks{\hspace{.5em}Now at Google Research.}\\\texttt{\small mmitchellai@google.com} \And Dirk Hovy\\University of Copenhagen\\\texttt{\small mail@dirkhovy.com}}
\date{}

\newcommand{\LR}{LR}
\newcommand{\STL}{STL}

\newcommand{\MTL}{MTL}

\newcommand{\removed}[1]{}

\usepackage{tikz,pgfplots,tikzscale}
\usetikzlibrary{shapes,arrows,shadows,fit,positioning}
\usetikzlibrary{external}

\usepackage{amsmath,bm,times}

\begin{document}

\pgfdeclarelayer{background}
\pgfdeclarelayer{foreground}
\pgfsetlayers{background,main,foreground}


\tikzstyle{inlayer}  = [draw, text width=10em, fill=gray!20, 
    minimum height=1.5em, rounded corners, text centered]
\tikzstyle{midlayer} = [draw, text width=6em, fill=gray!20, 
    minimum height=1.5em, rounded corners, text centered]
\tikzstyle{outlayer} = [draw, text width=4em, fill=gray!20, 
    minimum height=1.5em, rounded corners, text centered]
\tikzstyle{bubblebox} = [draw, text width=4em, fill=gray!20, 
    minimum height=1.5em, text centered]

\def\blockdist{2.3}
\def\edgedist{2.5}


\maketitle

\begin{abstract}

We introduce initial groundwork for estimating suicide risk and mental health in a deep learning framework. By modeling multiple conditions, the system learns to make predictions about suicide risk and mental health at a low false positive rate.  Conditions are modeled as tasks in a multi-task learning (MTL) framework, with gender prediction as an additional auxiliary task. We demonstrate the effectiveness of multi-task learning by comparison to a well-tuned single-task baseline with the same number of parameters. Our best MTL model predicts potential suicide attempt, as well as the presence of atypical mental health, with AUC $>$ 0.8.  We also find additional large improvements using multi-task learning on mental health tasks with limited training data. 
\end{abstract}


\section{Introduction}
Suicide is one of the leading causes of death worldwide, and over 90\% of individuals who die by suicide experience mental health conditions.\footnote{\small{https://www.nami.org/Learn-More/Mental-Health-Conditions/Related-Conditions/Suicide\#sthash.dMAhrKTU.dpuf}} However, detecting the risk of suicide, as well as monitoring the effects of related mental health conditions, is challenging.  Traditional methods rely on both self-reports and impressions formed during short sessions with a clinical expert, but it is often unclear when suicide is a risk in particular.\footnote{Communication with clinicians at the 2016 JSALT workshop \cite{ws16ehr}.} Consequently, conditions leading to preventable suicides are often not adequately addressed.


Automated monitoring and risk assessment of patients' language has the potential to complement traditional assessment methods, providing objective measurements to motivate further care and additional support for people with difficulties related to mental health. This paves the way towards verifying the need for additional care with insurance coverage, for example, as well as offering direct benefits to clinicians and patients.

We explore some of the possibilities in the deep learning and mental health space using {\it written social media text} that people with different mental health conditions are already producing.  Uncovering methods that work with such text provides the opportunity to help people with different mental health conditions by leveraging a task they are already participating in.

Social media text carries implicit information about the author, which has been modeled in natural language processing (NLP) to predict author characteristics such as \emph{age} \cite{Goswami:ea:2009stylometric,Rosenthal:ea:2011age,Nguyen:ea:14}, 
\emph{gender} \cite{Sarawgi:ea:2011gender,Ciot:ea:2013,Liu:ea:2013,Volkova:ea:15inferring,hovy2015demographic}, \emph{personality} \cite{schwartz2013toward,volkova2014inferring,plank2015personality,park:ea:2015,preotiuc-pietro:ea:2015personality}, 
and \emph{occupation} \cite{preotiuc2015analysis}. 
Similar text signals have been effectively used to predict mental health conditions such as \emph{depression} \cite{dechoudhury2013predicting,W15-1204,W14-3214}, \emph{suicidal ideation} \cite{W16-0311,Y15-1064}, \emph{schizophrenia} \cite{W15-1202} or \emph{post-traumatic stress disorder (PTSD)} \cite{W15-1206}.

However, these studies typically model each condition in isolation, which misses the opportunity to model coinciding influence factors.  Tasks with underlying commonalities (e.g., part-of-speech tagging, parsing, and NER) have been shown to benefit from multi-task learning (\MTL{}), as the learning implicitly leverages interactions between them ~\cite{caruana1993multitask,sutton:ea:2007,rush2010dual,collobert:ea:2011,sogaard2016deep}. Suicide risk and related mental health conditions are therefore good candidates for modeling in a multi-task framework.  

In this paper, we propose multi-task learning for detecting suicide risk and mental health conditions. The tasks of our model include \emph{neuroatypicality} (i.e., atypical mental health) and \emph{suicide attempt}, as well as the related mental health conditions of \emph{anxiety}, \emph{depression}, \emph{eating disorder}, \emph{panic attacks}, \emph{schizophrenia}, \emph{bipolar disorder}, and \emph{post-traumatic stress disorder (PTSD)}, and we explore the effect of task selection on model performance. We additionally include the effect of modeling \emph{gender}, which has been shown to improve accuracy in tasks using social media text \cite{Volkova:ea:2013exploring,hovy2015demographic}.

Predicting suicide risk and several mental health conditions jointly opens the possibility for the model to leverage a shared representation for conditions that frequently occur together, a phenomenon known as \emph{comorbidity}. Further including gender reflects the fact that gender differences are found in the patterns of mental health \cite{WHO}, which may help to sharpen the model.  The \MTL{} framework we propose allows such shared information across predictions and enables the inclusion of several loss functions with a common shared underlying representation.  This approach is flexible enough to extend to factors other than the ones shown here, provided suitable data.

We find that choosing tasks that are prerequisites or related to the main task is critical for learning a strong model, similar to \newcite{caruana1996algorithms}.  We further find that modeling gender improves accuracy across a variety of conditions, including suicide risk. The best-performing model from our experiments demonstrates that multi-task learning is a promising new direction in automated assessment of mental health and suicide risk, with possible application to the clinical domain.

\paragraph{Our contributions}
\begin{enumerate}
\itemsep-0.5em
\item We demonstrate the utility of \MTL{} in predicting mental health conditions from social user text -- a notoriously difficult task \cite{coppersmith2015adhd,W15-1204} -- with potential application to detecting suicide risk. \NoteMM{Anyone else we can cite to establish the notoriety?}
\item We explore the influence of task selection on prediction performance, including the effect of gender.
\item We show how to model tasks with a large number of positive examples to improve the prediction accuracy of tasks with a small number of positive examples.
\item We compare the MTL model against a single-task model with the same number of parameters, which directly evaluates the multi-task learning approach.
\item The proposed MTL model increases the True Positive Rate at 10\% false alarms by up to 9.7\% absolute (for anxiety), a result with direct impact for clinical applications.
\end{enumerate}

\NoteMM{These are results for *directly porting* to clinical.  Keep separate from *research findings* to move forward in this vein.}


\section{Ethical Considerations}
As with any author-attribute detection, there is the danger of abusing the model to single out people (\emph{overgeneralization}, see \newcite{hovysocial}). We are aware of this danger, and sought to minimize the risk. For this reason, we don't provide a selection of features or representative examples. The experiments in this paper were performed with a clinical application in mind, and use carefully matched (but anonymized) data, so the distribution is not representative of the population as a whole. The results of this paper should therefore \emph{not} be interpreted as a means to assess mental health conditions in social media in general, but as a test for the applicability of MTL in a well-defined clinical setting.

\section{Model Architecture}
A neural multi-task architecture opens the possibility of leveraging commonalities and differences between mental conditions.  Previous work \cite{collobert:ea:2011,caruana1996algorithms,caruana1993multitask} has indicated that such an architecture allows for sharing parameters across tasks, and can be beneficial when there is 
varying degrees of annotation across tasks.\footnote{We also experimented with a graphical model architecture, but found that it did not scale as well and provided less traction.} 
This makes MTL particularly compelling in light of mental health comorbidity, and given that different conditions have different amounts of associated data. 

Previous \MTL{} approaches 
have shown considerable improvements over single task models, and the arguments are convincing: Predicting multiple related tasks should allow us to exploit any correlations between the predictions.  However, in much of this work, an \MTL{} model is only one possible explanation for improved accuracy.  Another more salient factor has frequently been overlooked: The difference in the expressivity of the model class, i.e., neural architectures vs.~discriminative or generative models, and critically, differences in the number of parameters for comparable models.  Some comparisons might therefore have inadvertently compared apples to oranges.
\NoteAB{Do we still want to emphasize this?}
\NoteMM{Definitely, but no need to double down -- it is one contribution, not main point of paper (right?)}
\NoteDH{Yes, it should still be in there, because it's often overlooked, but it's not the main point anymore}

In the interest of examining the effect of multi-task learning specifically, we compare the multi-task predictions to models with equal expressivity. We evaluate the performance of a standard logistic regression model (a standard approach to text-classification problems), a multilayer perceptron single-task learning (\STL) model, and a neural \MTL{} model, the latter two with equal numbers of parameters.
This ensures a fair comparison by isolating the unique properties of \MTL{} from the dimensionality-reduction aspects of deep architectures in general.

\begin{figure}
    \begin{center}
	\includegraphics[trim=0cm 0.5cm 2.0cm 2.0cm, width=0.8\textwidth]{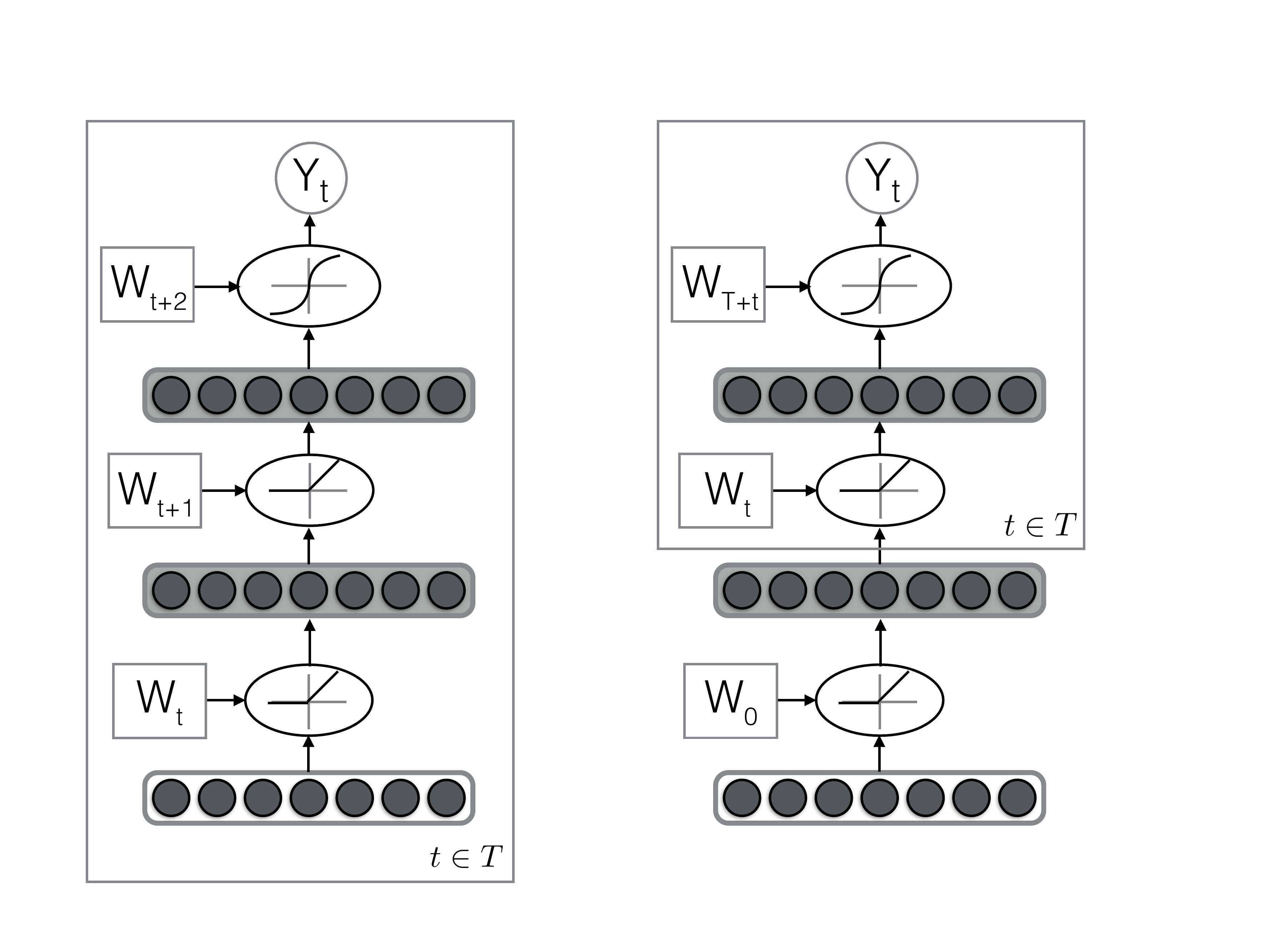} \\
    {\bf Single-task} \hspace{2.75em} {\bf Multi-task}
    \end{center}
	\caption{STL model in plate notation (left): weights trained independently for each task $t$ (e.g., anxiety, depression) of the $T$ tasks.  MTL model (right): shared weights trained jointly for all tasks, with task-specific hidden layers.  Curves in ovals represent the type of activation used at each layer (rectified linear unit or sigmoid). Hidden layers are shaded.}
\label{fig:model}
\end{figure}

The neural models we evaluate come in two forms.  The first, depicted in plate notation on the left in Figure \ref{fig:model}, are the \STL{} models.  These are feedforward networks with two hidden layers, trained independently to predict each task.  On the right in Figure \ref{fig:model} is the \MTL{} model, where the first hidden layer from the bottom is shared between all tasks.  An additional per-task hidden layer is used to give the model flexibility to map from the task-agnostic representation to a task-specific one.  Each hidden layer uses a rectified linear unit as non-linearity.  The output layer uses a logistic non-linearity, since all tasks are binary predictions. 
The \MTL{} model can easily be extended to a stack of shared hidden layers,
allowing for a more complicated mapping from input to shared space.\footnote{We tried training a 4-shared-layer \MTL{} model to predict targets on a separate dataset, but did not see any gains over the standard 1-shared-layer \MTL{} model in our application.}

As noted in \newcite{collobert:ea:2011}, MTL benefits from mini-batch training, which both allows optimization to jump out of poor local optima, and more stochastic gradient steps in a fixed amount of time \cite{bottou2012}. 
We create mini-batches by sampling from the users in our data, where each user has some subset of the conditions we are trying to predict, and may or may not be annotated with gender.  At each mini-batch gradient step, we update weights for all tasks.  This not only allows for randomization and faster convergence, it also provides a speed-up over the individual selection process reported in earlier work \cite{collobert:ea:2011}.

Another advantage of this setup is that we do not need complete information for every instance: Learning can proceed with asynchronous updates, dependent on what the data in each batch has been annotated for, while sharing representations throughout.  This effectively learns a joint model with a common representation for several different tasks, allowing the use of several ``disjoint'' data sets, some with limited annotated instances.

\begin{figure*}[ht!]
  \begin{center}
		\includegraphics[width=0.3\textwidth,trim={0 12.1cm 0.1cm 0},clip]{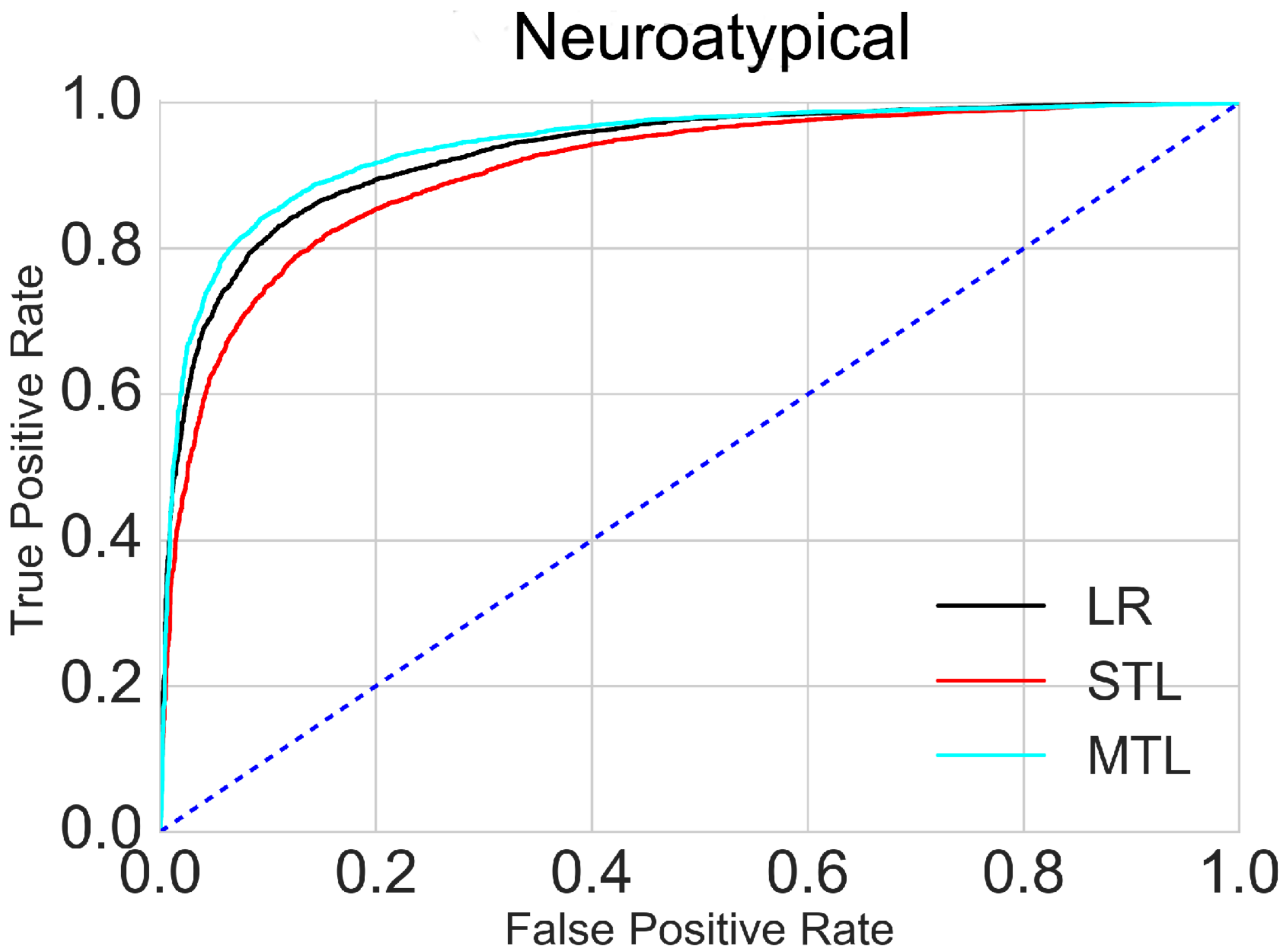} \hspace{1em} 		\includegraphics[width=0.3\textwidth,trim={0 12.25cm 0.1cm 0},clip]{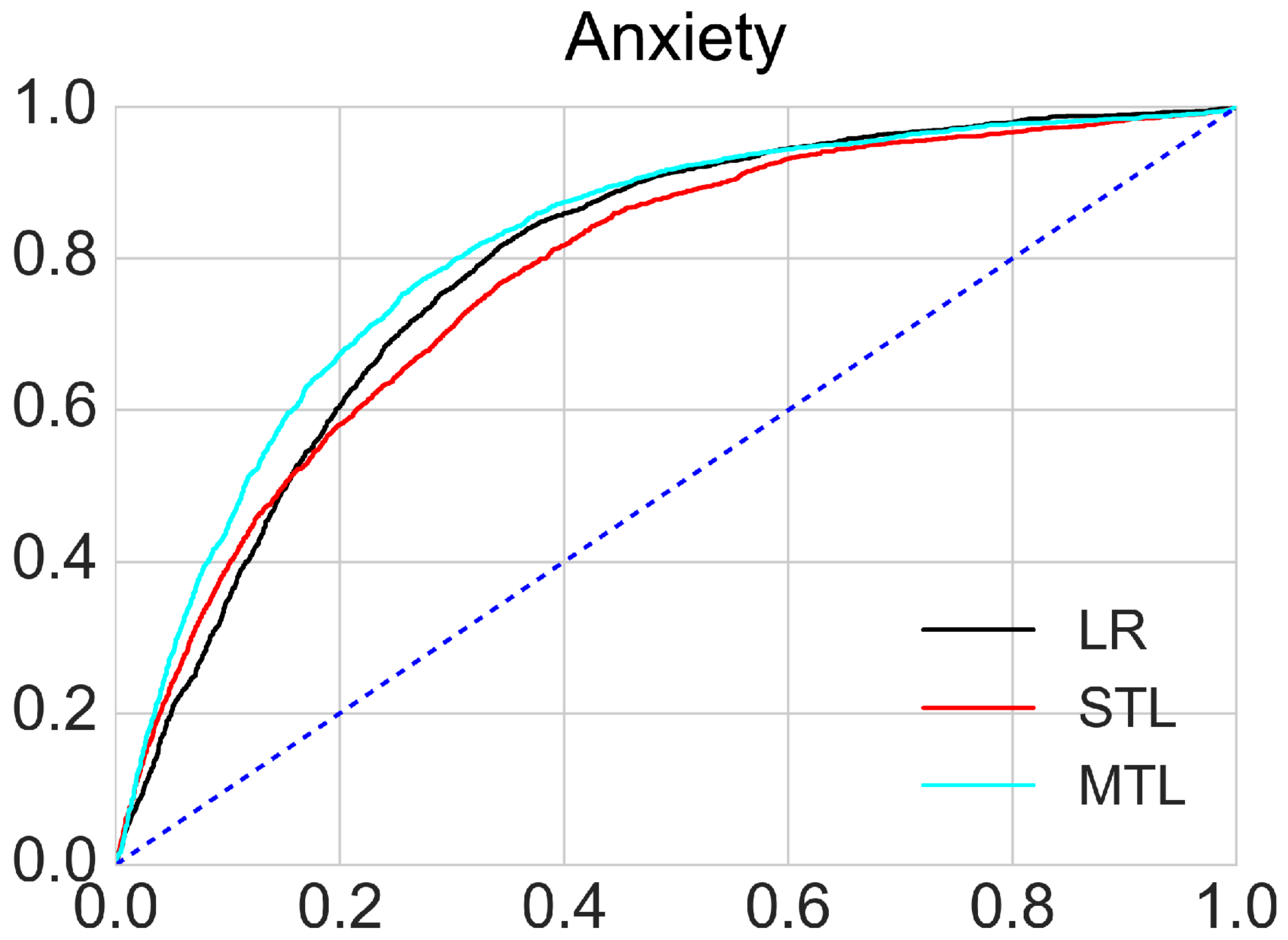}
\hspace{1em}		\includegraphics[width=0.3\textwidth,trim={0 12.25cm 0.1cm 0},clip]{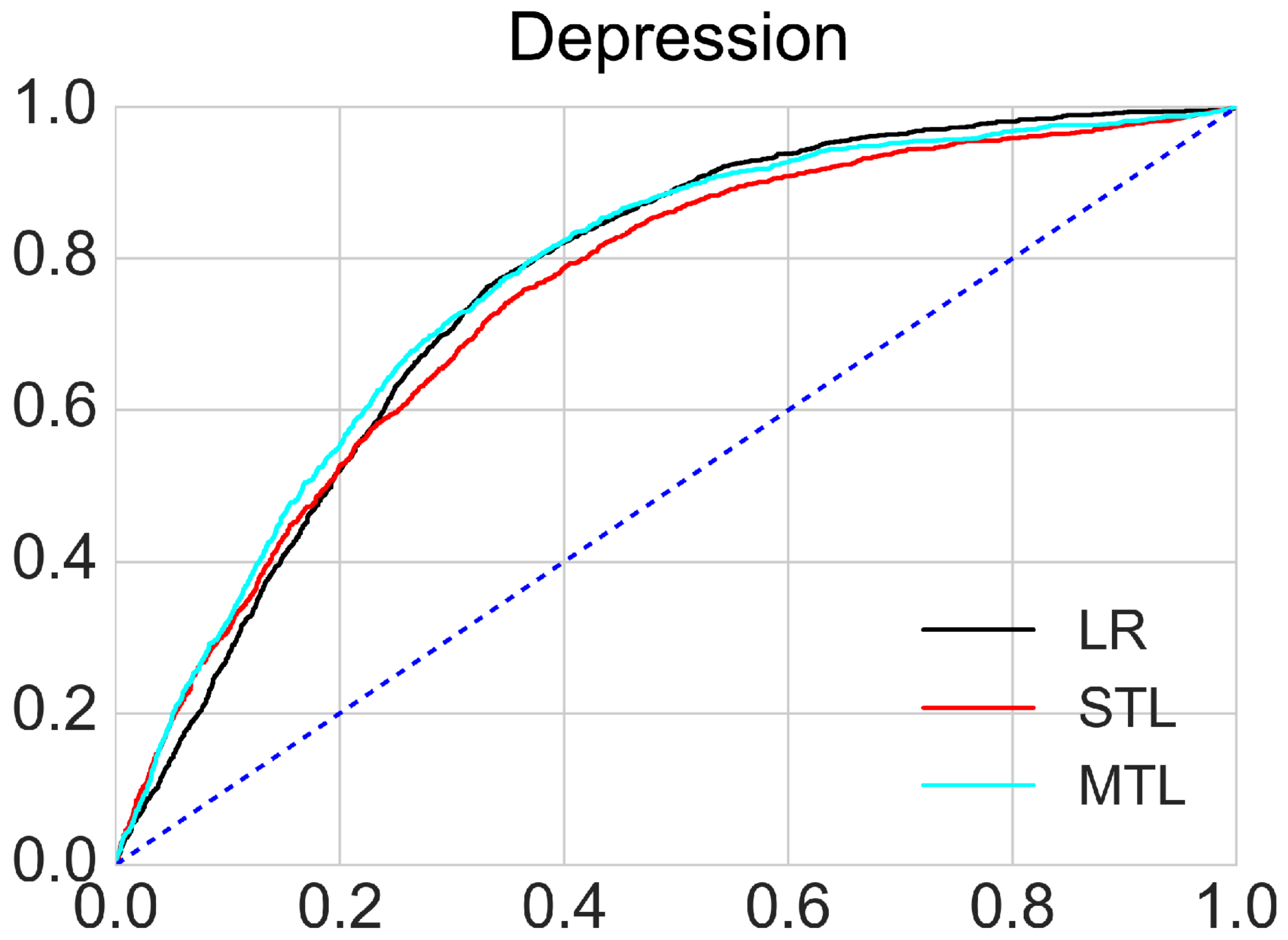}\\\vspace{1em}
		\includegraphics[width=0.3\textwidth,trim={0 12.25cm 0.1cm 0},clip]{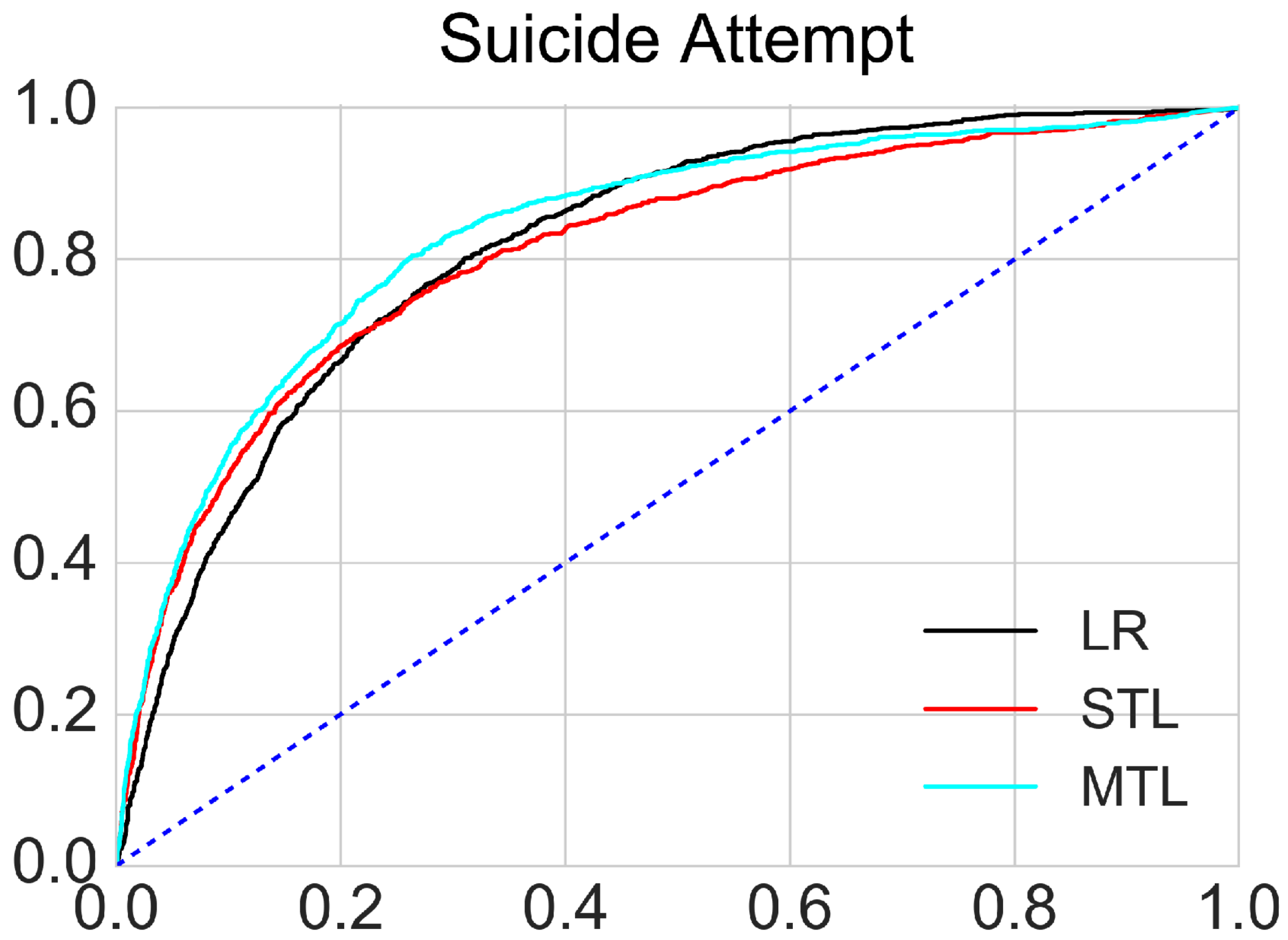}
        \hspace{1em}
		\includegraphics[width=0.3\textwidth,trim={0 12.25cm 0.1cm 0},clip]{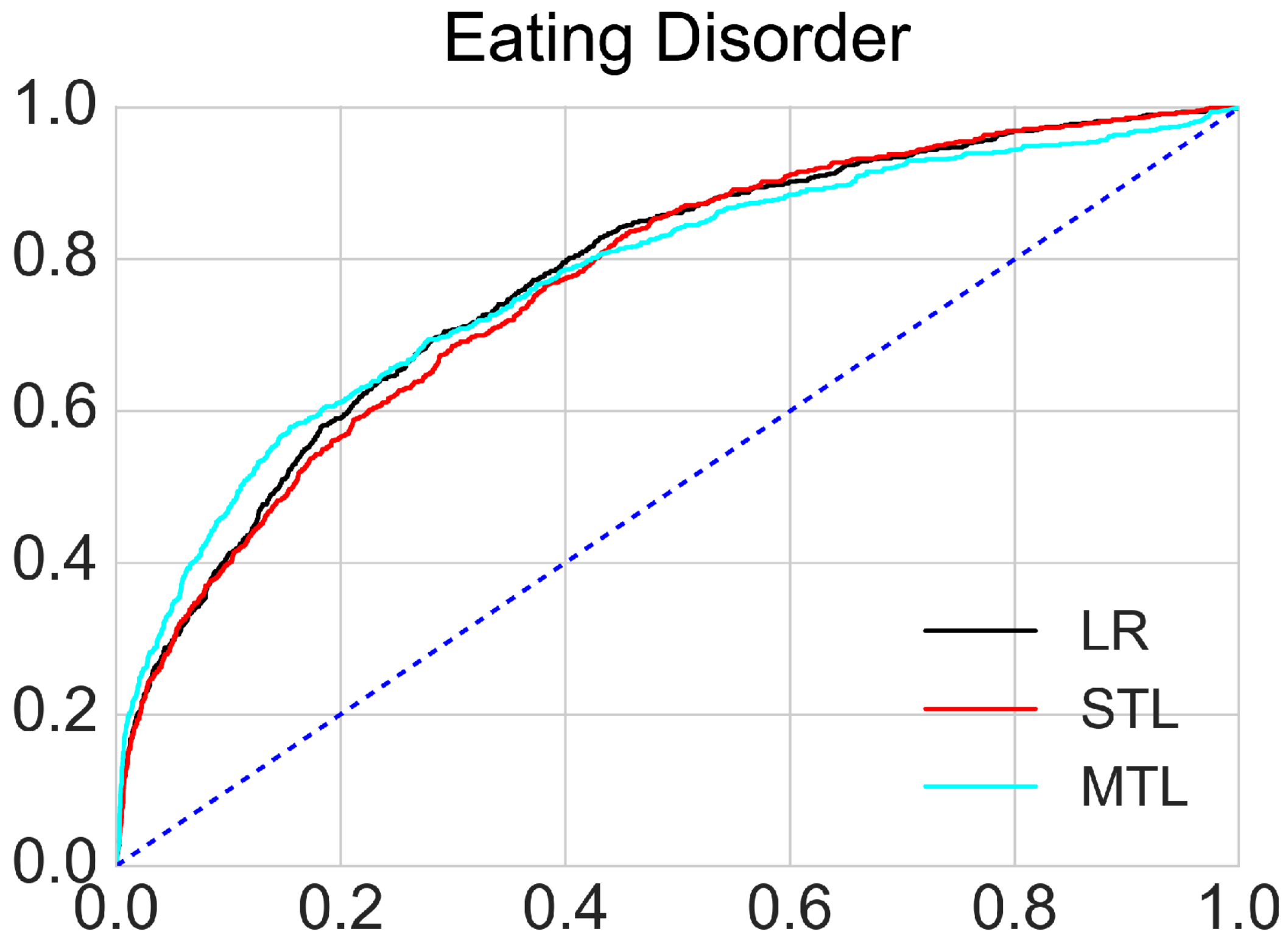}
        \hspace{1em}
		\includegraphics[width=0.3\textwidth,trim={0 12.25cm 0.1cm 0},clip]{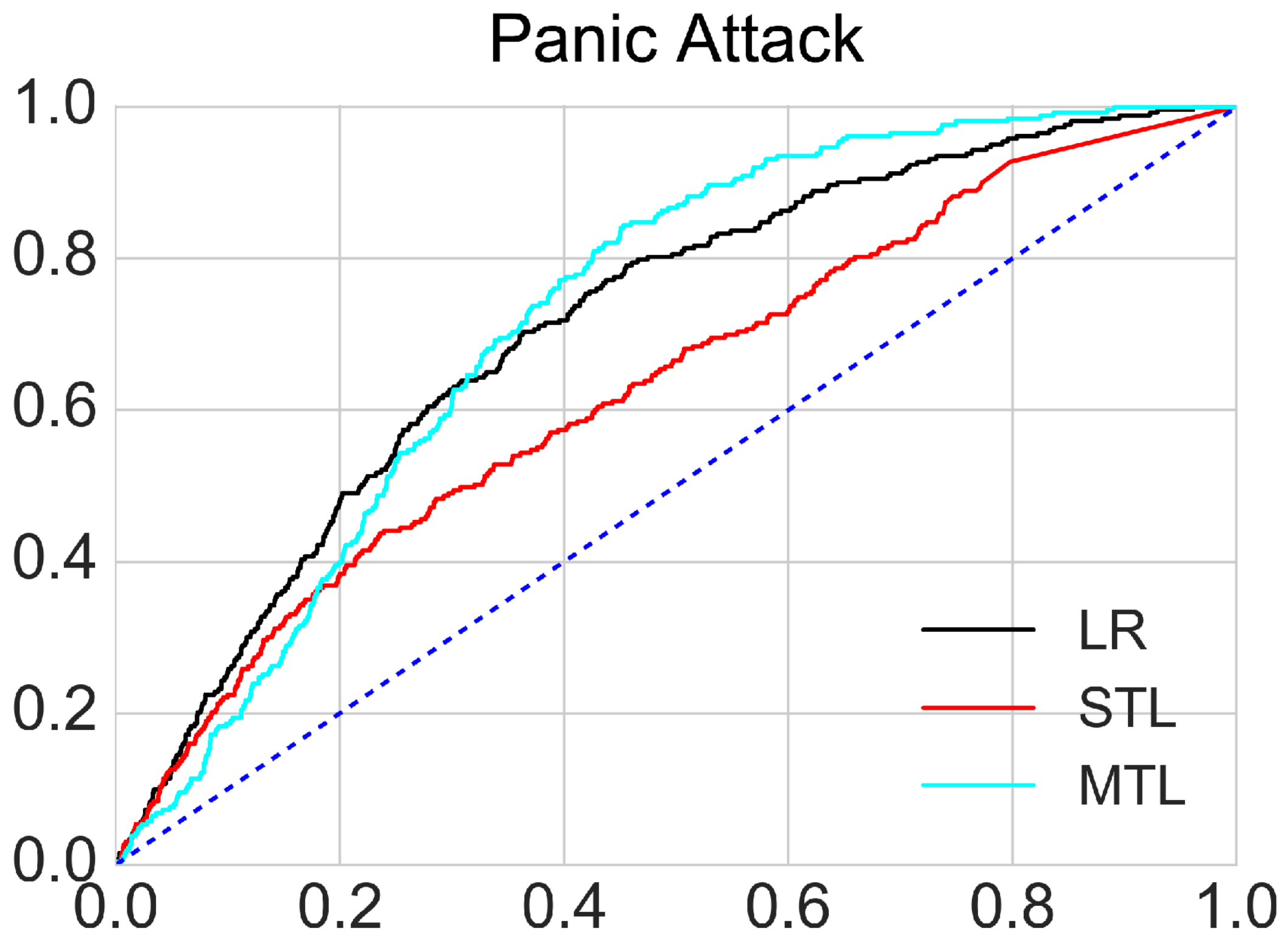}\\\vspace{1em}
		\includegraphics[width=0.3\textwidth,trim={0 12.25cm 0.1cm 0},clip]{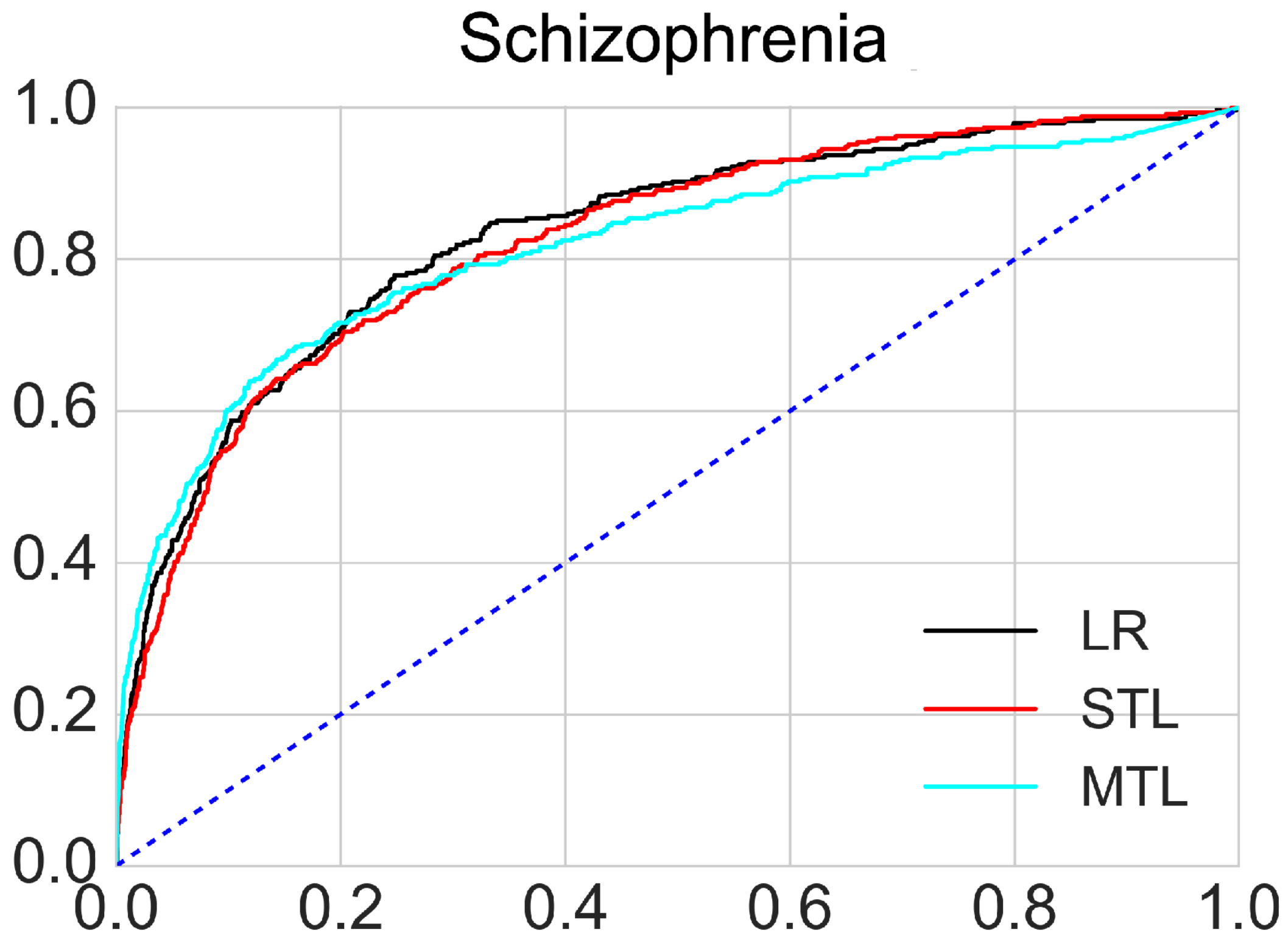}
        \hspace{1em}
		\includegraphics[width=0.3\textwidth,trim={0 12.25cm 0.1cm 0},clip]{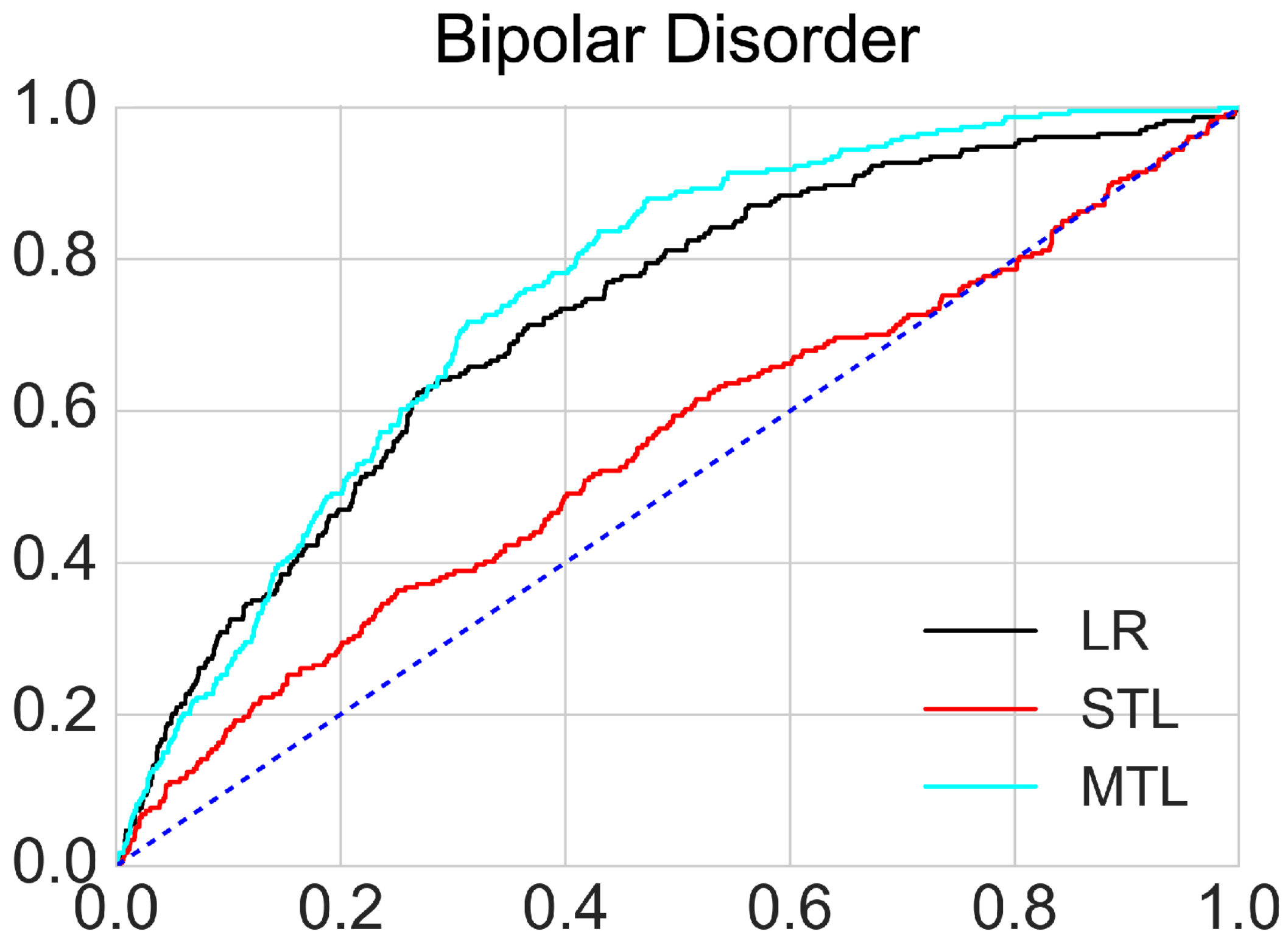}
        \hspace{1em}
		\includegraphics[width=0.3\textwidth,trim={0 12.25cm 0.1cm 0},clip]{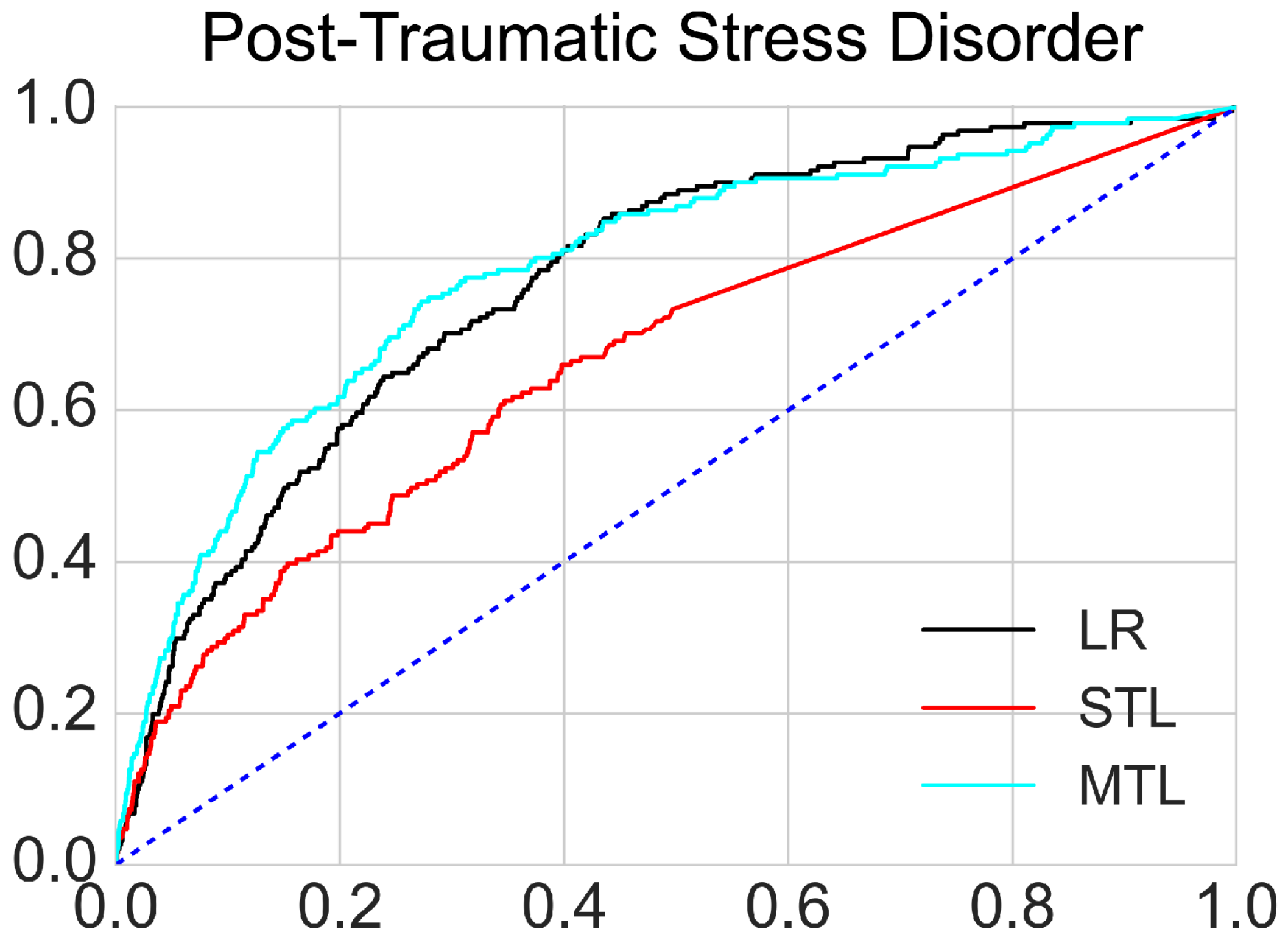}
		\captionof{figure}{ROC curves for predicting each condition. The precision (diagnosed, correctly labeled) is on the $y$-axis, while the proportion of false alarms (control users mislabeled as diagnosed) is on the $x$-axis. Chance performance is indicated by the dotted diagonal line. \label{pic:roc_curves}}
	\end{center}
\end{figure*}

\paragraph{Optimization and Model Selection}
Even in a relatively simple neural model, there are a number of hyperparameters that can (and have to) be tuned to achieve good performance. We perform a line search for every model we use, sweeping over L$_2$ regularization and hidden layer width.  We select the best model based on the development loss. Figure \ref{pic:roc_curves} shows the performance on the corresponding test sets (plot smoothed by rolling mean of 10 for visibility).

In our experiments, we sweep over the L2 regularization constant applied to all weights in $\{10^{-4}, 10^{-3}, 10^{-2}, 0.1, 0.5, 1.0, 5.0, 10.0\}$, and hidden layer width (same for all layers in the network) in $\{16, 32, 64, 128, 256, 512, 1024, 2048\}$.  We fix the mini-batch size to 256, and 0.05 dropout on the input layer.  Choosing a small mini-batch size and the model with lowest development loss helps to account for overfitting.


We train each model for 5,000 iterations, jointly updating all weights in our models.  After this initial joint training, we select each task separately, and only update the task-specific layers of weights independently for another 1,000 iterations (selecting the set of weights achieving lowest development loss for each task individually). Weights are updated using mini-batch Adagrad \cite{duchi2011} -- this converges more quickly than other optimization schemes we experimented with.  We evaluate the tuning loss every 10 epochs, and select the model with the lowest tuning loss.

\section{Data}

\renewcommand{\arraystretch}{1.25}
\begin{table*}
\small
\begin{tabular}{r||r|r|r|r|r|r|r|r|r||r|r|}
\cline{2-12}
& \rot{\sc neurotypical}	 & \rot{\sc anxiety}	 & \rot{\sc depression}	 & \rot{\sc suicide attempt}	 & \rot{\sc eating}	 & \rot{\sc schizophrenia}	 & \rot{\sc panic}	 & \rot{\sc PTSD}	 & \rot{\sc bipolar}  & \rot{\sc labeled male} & \rot{\sc labeled female} \\\hline
{\sc neurotypical}	 & 4820	 & 	\multicolumn{8}{c||}{~}  & - & - \\\cline{2-3}\cline{11-12}
{\sc anxiety}	 & 0	 & 2407	 	 & 	\multicolumn{7}{c||}{~}& 47 & 184\\\cline{2-4}\cline{11-12}
{\sc depression}	 & 0	 & 1148	 & 1400	  & 	\multicolumn{6}{c||}{~} & 54 & 158\\\cline{2-5}\cline{11-12}
{\sc suicide attempt}	 & 0	 & 45	 & 149	 & 1208	& 	 \multicolumn{5}{c||}{~}& 186 & 532\\\cline{2-6}\cline{11-12}
{\sc eating}	 & 0	 & 64	 & 133	 & 45	 & 749	 & 	 \multicolumn{4}{c||}{~}& 6 & 85 \\\cline{2-7}\cline{11-12}
{\sc schizophrenia}	 & 0	 & 18	 & 41	 & 2	 & 8	 & 349	 & \multicolumn{3}{c||}{~}& 2 & 4 \\\cline{2-8}\cline{11-12}
{\sc panic}	 & 0	 & 136	 & 73	 & 4	 & 2	 & 4	 & 263	 & 	\multicolumn{2}{c||}{~} &2 & 18 \\\cline{2-9}\cline{11-12}
{\sc PTSD}	 & 0	 & 143	 & 96	 & 14	 & 16	 & 14	 & 22	 & 191		&& 8 & 26 \\\cline{2-10}\cline{11-12}
{\sc bipolar} & 0	 & 149	 & 120	 & 22	 & 22	 & 49	 & 14	 & 25	 & 234	& 10 & 39 \\\cline{2-12}
\end{tabular}
\caption{Frequency and comorbidity across mental health conditions.}
\label{table:data}
\end{table*}

We train our models on a union of multiple Twitter user datasets: 1) users identified as having anxiety, bipolar disorder, depression, panic disorder, eating disorder, PTSD, or schizophrenia \cite{coppersmith2015adhd}, 2) those who had attempted suicide \cite{coppersmith2015quantifying}, and 3) those identified as having either depression or PTSD from the 2015 Computational Linguistics and Clinical Psychology Workshop shared task \cite{W15-1204}, along with neurotypical gender-matched controls (Twitter users not identified as having a mental condition).  Users were identified as having one of these conditions if they stated explicitly they were diagnosed with this condition on Twitter (verified by a human annotator), and the data was pre-processed to remove direction indications of the condition.  For a subset of 1,101 users, we also manually-annotate gender.  The final dataset contains 9,611 users in total, with an average of 3521 tweets per user.  The number of users with each condition is included in Table \ref{table:data}. Users in this joined dataset may be tagged with multiple conditions, thus the counts in this table do not sum to the total number of users.

We use the entire Twitter history of each user as input to the model, and split it into character 1-to-5-grams, which have been shown to capture more information than words for many Twitter text classification tasks \cite{mcnamee2004character,coppersmith2015adhd}. We compute the relative frequency of the 5,000 most frequent $n$-gram features for $n \in \{1,2,3,4,5\}$ in our data, and then feed this as input to all models.  This input representation is common to all models, allowing for fair comparison.


\section{Experiments}

Our task is to predict suicide attempt and mental conditions for each of the users in these data. 
We evaluate three classes of models: baseline logistic regression over character $n$-gram features (\LR), feed-forward multilayer perceptrons trained to predict each task separately (\STL), and feed-forward multi-task models trained to predict a set of conditions simultaneously (\MTL).  We experiment with a feed-forward network against independent logistic regression models as a way to directly test the hypothesis that MTL may work well in this domain.

We also perform ablation experiments to see which subsets of tasks help us learn an \MTL{} model that predicts a particular mental condition best.  For all experiments, data were divided into five equal-sized folds, three for training, one for tuning, and one for testing (we report the performance on this).  






All our models are implemented in Keras\footnote{\url{http://keras.io/}} with Theano backend and GPU support.  We train the models for a total of up to 15,000 epochs, using mini-batches of 256 instances. Training time on all five training folds ranged from one to eight hours on a machine with Tesla K40M.

\paragraph{Evaluation Setup} 
We compare the accuracy of each model at predicting each task separately. 

In clinical settings, we are interested in minimizing the number of false positives, i.e., incorrect diagnoses, which can cause undue stress to the patient.  We are thus interested in bounding this quantity. To evaluate the performance, we plot the false positive rate (FPR) against the true positive rate (TPR). This gives us a receiver operating characteristics (ROC) curve, allowing us to inspect the performance of each model on a specific task at any level of FPR.

While the ROC gives us a sense of how well a model performs at a fixed true positive rate, it makes it difficult to compare the individual tasks at a low false positive rate, which is also important for clinical application. We therefore report two more measures: the area under the ROC curve (AUC) and TPR performance at FPR=0.1 (TPR@FPR=0.1).  We do not compare our models to a majority baseline model, since this model would achieve an expected AUC of 0.5 for all tasks, and F-score and TPR@FPR=0.1 of 0 for all mental conditions -- users exhibiting a condition are the minority, meaning a majority baseline classifier would achieve zero recall.

\section{Results}

\begin{figure}[ht!]
	\begin{center}
		\includegraphics[width=\textwidth]{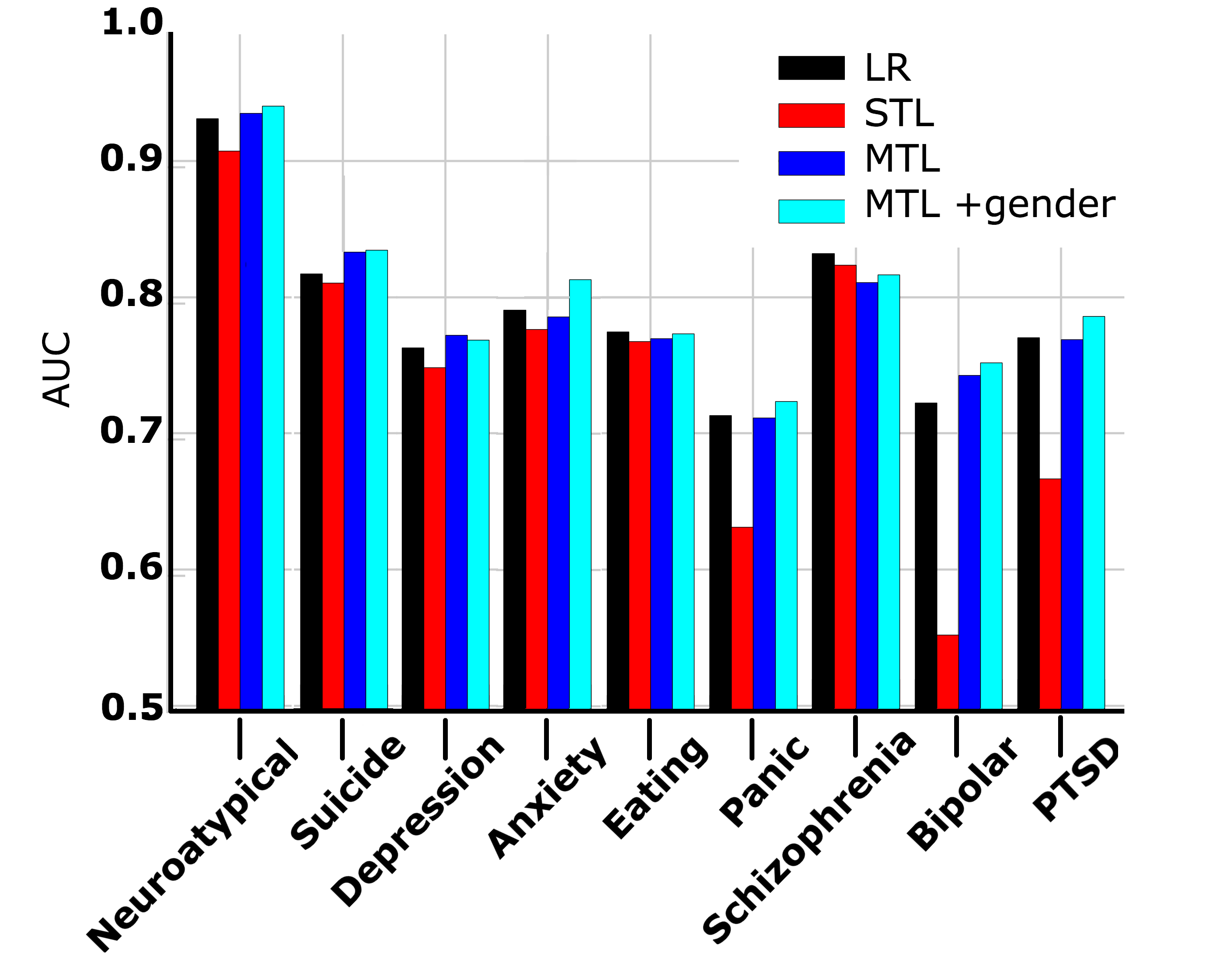}
		\captionof{figure}{AUC for different main tasks \label{pic:auc}}
	\end{center}
\end{figure}
\begin{figure}[ht!]
	\begin{center}
		\includegraphics[width=\textwidth]{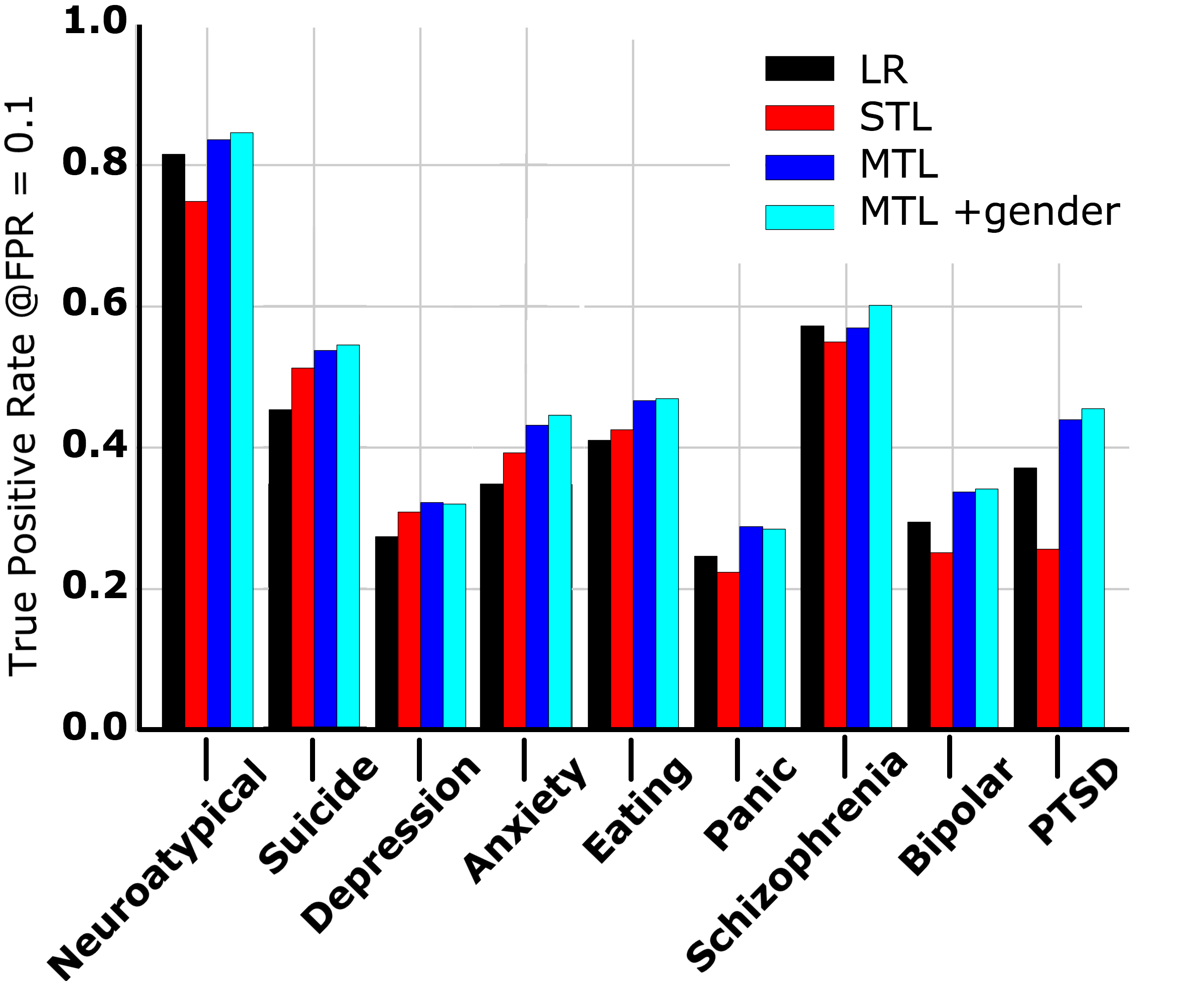}
		\captionof{figure}{TPR at 0.10 FPR for different main tasks \label{pic:tpr}}
	\end{center}
\end{figure}
\begin{figure*}[ht!]
  \begin{center}
		\includegraphics[width=0.3\textwidth,trim={0 12.25cm 0.1cm 0},clip]{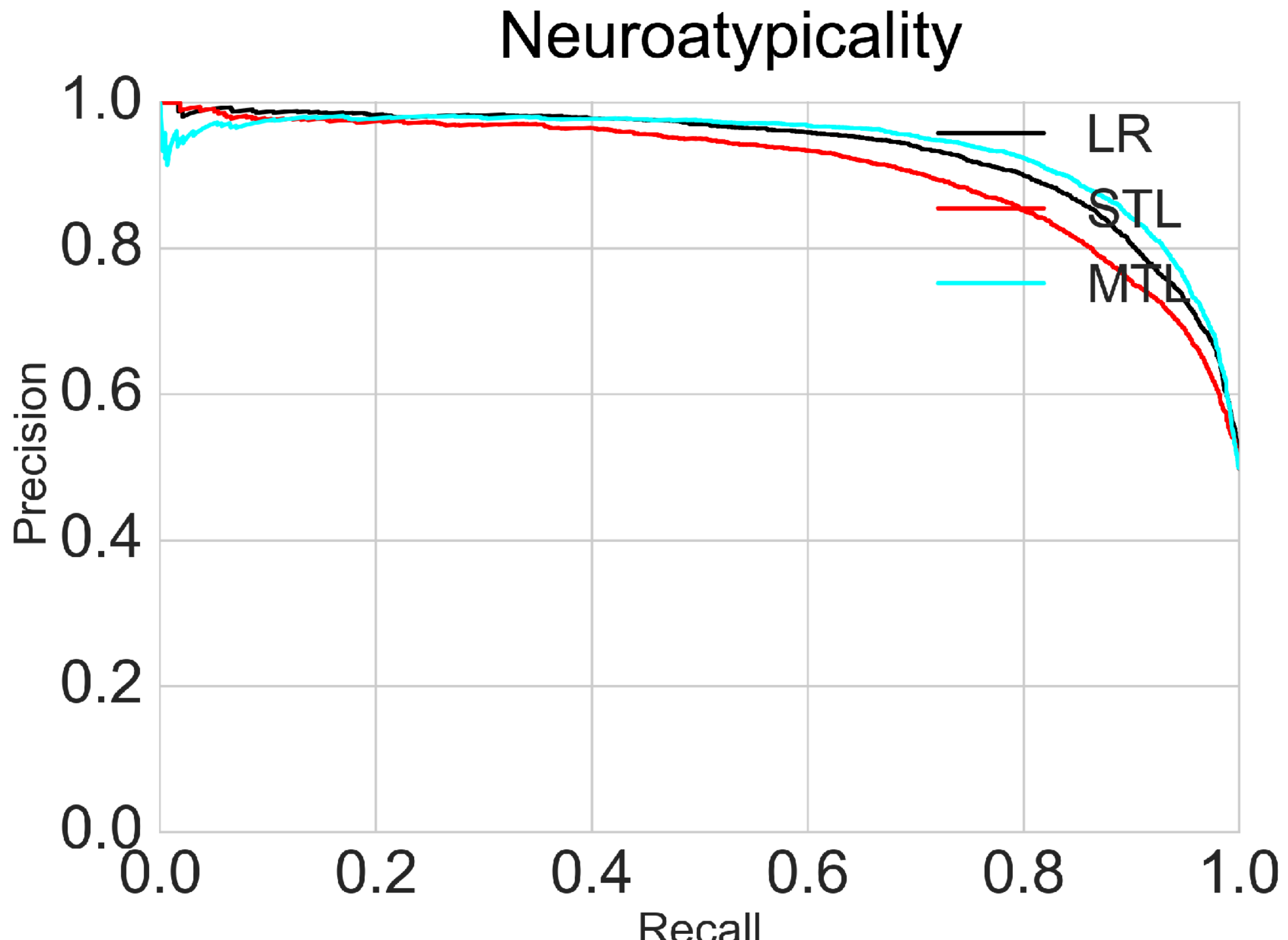}
\hspace{1em}	\includegraphics[width=0.3\textwidth,trim={0 12.25cm 0.1cm 0},clip]{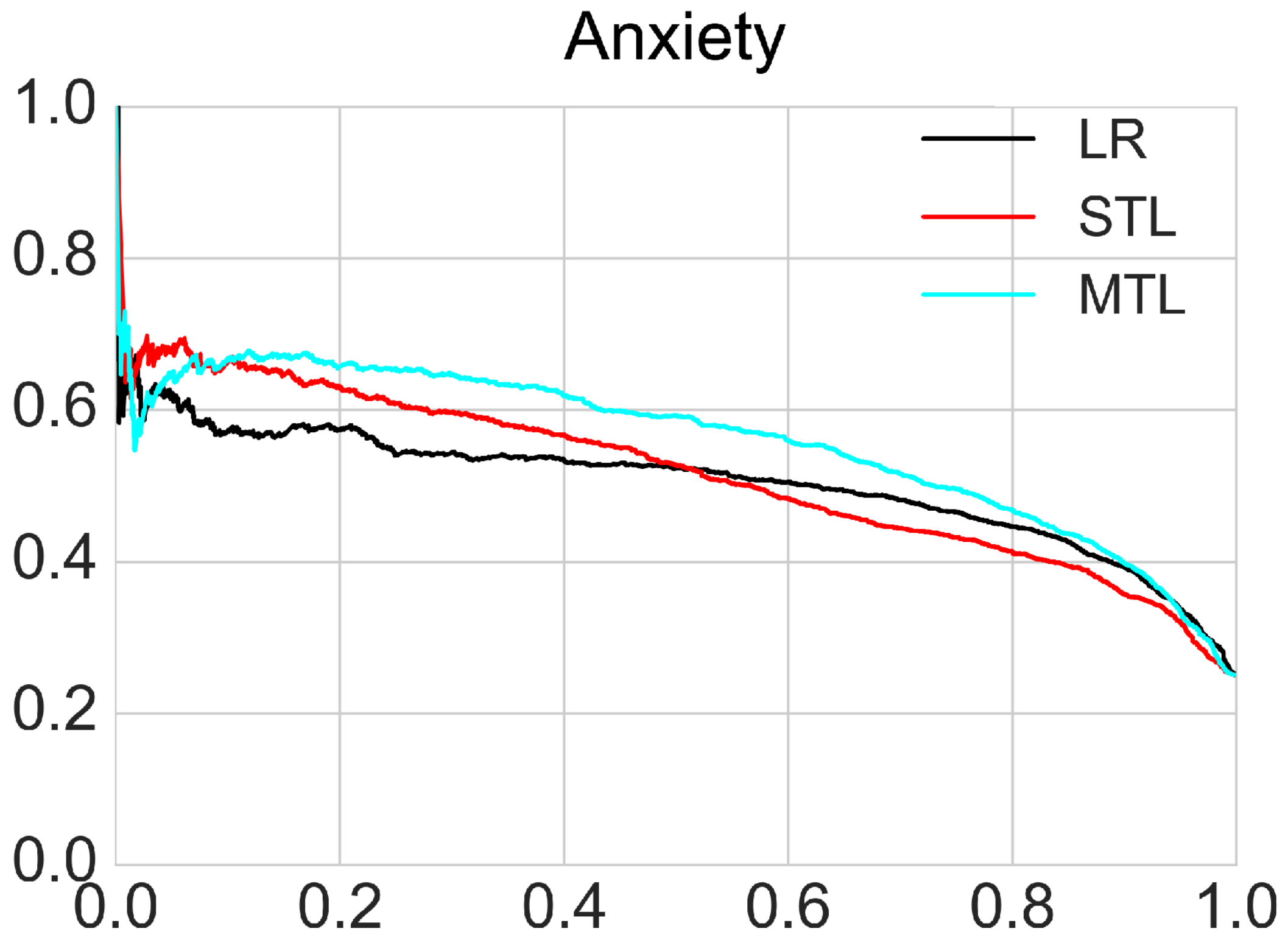}
\hspace{1em}	\includegraphics[width=0.3\textwidth,trim={0 12.25cm 0.1cm 0},clip]{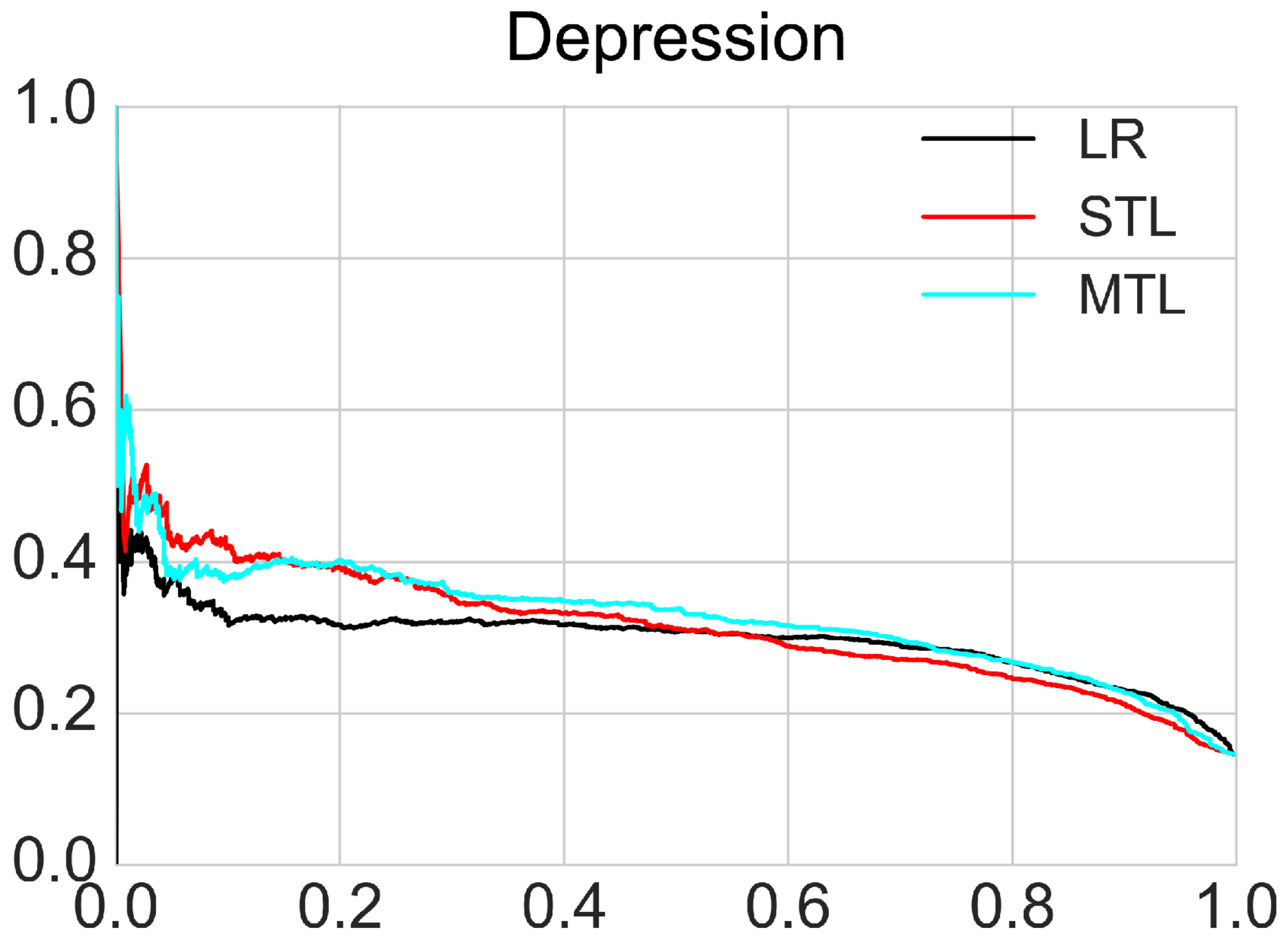} \\
\vspace{1em}
\includegraphics[width=0.3\textwidth,trim={0 12.25cm 0.1cm 0},clip]{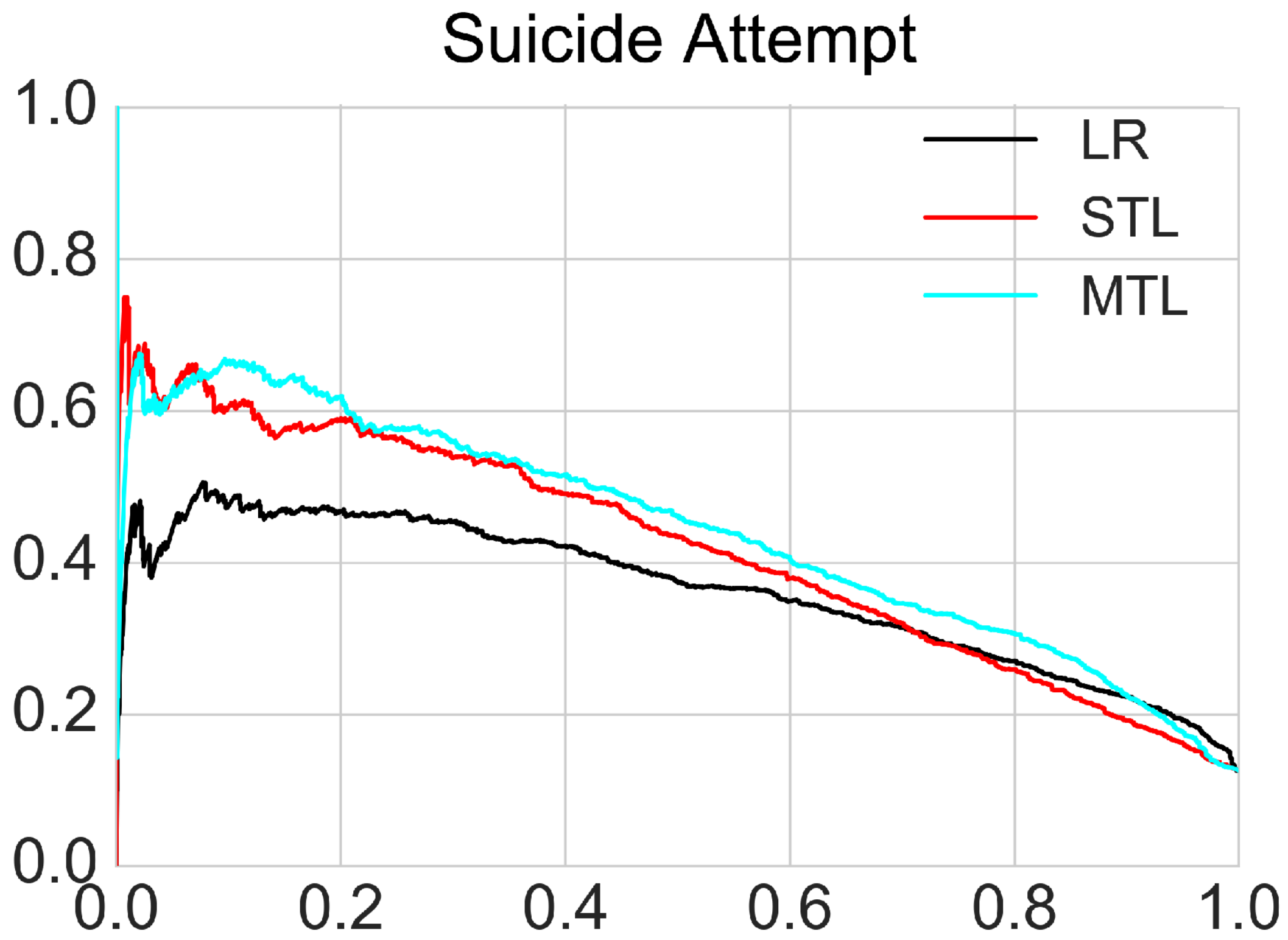}
	\hspace{1em}	\includegraphics[width=0.3\textwidth,trim={0 12.25cm 0.1cm 0},clip]{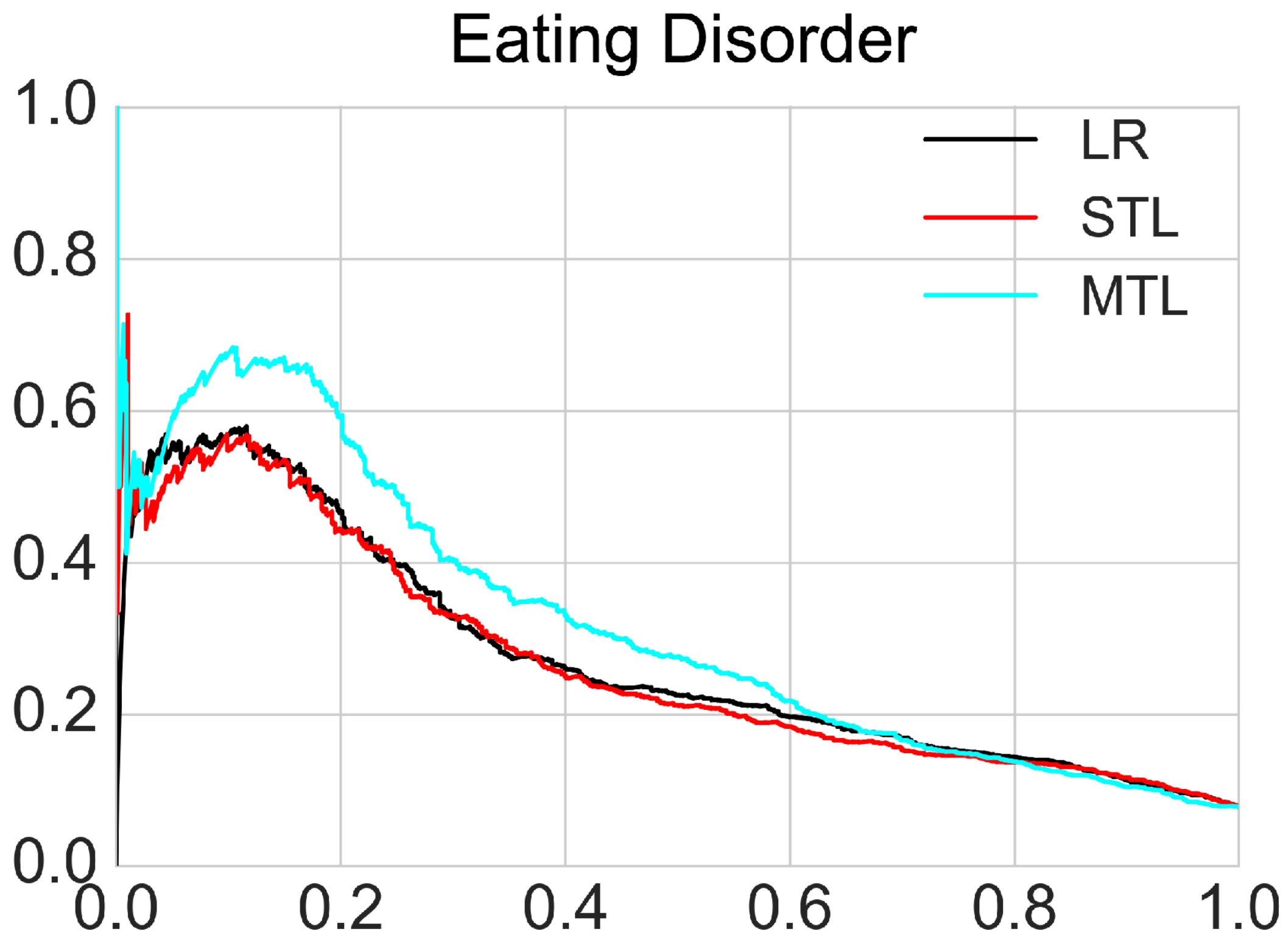}
	\hspace{1em}	\includegraphics[width=0.3\textwidth,trim={0 12.25cm 0.1cm 0},clip]{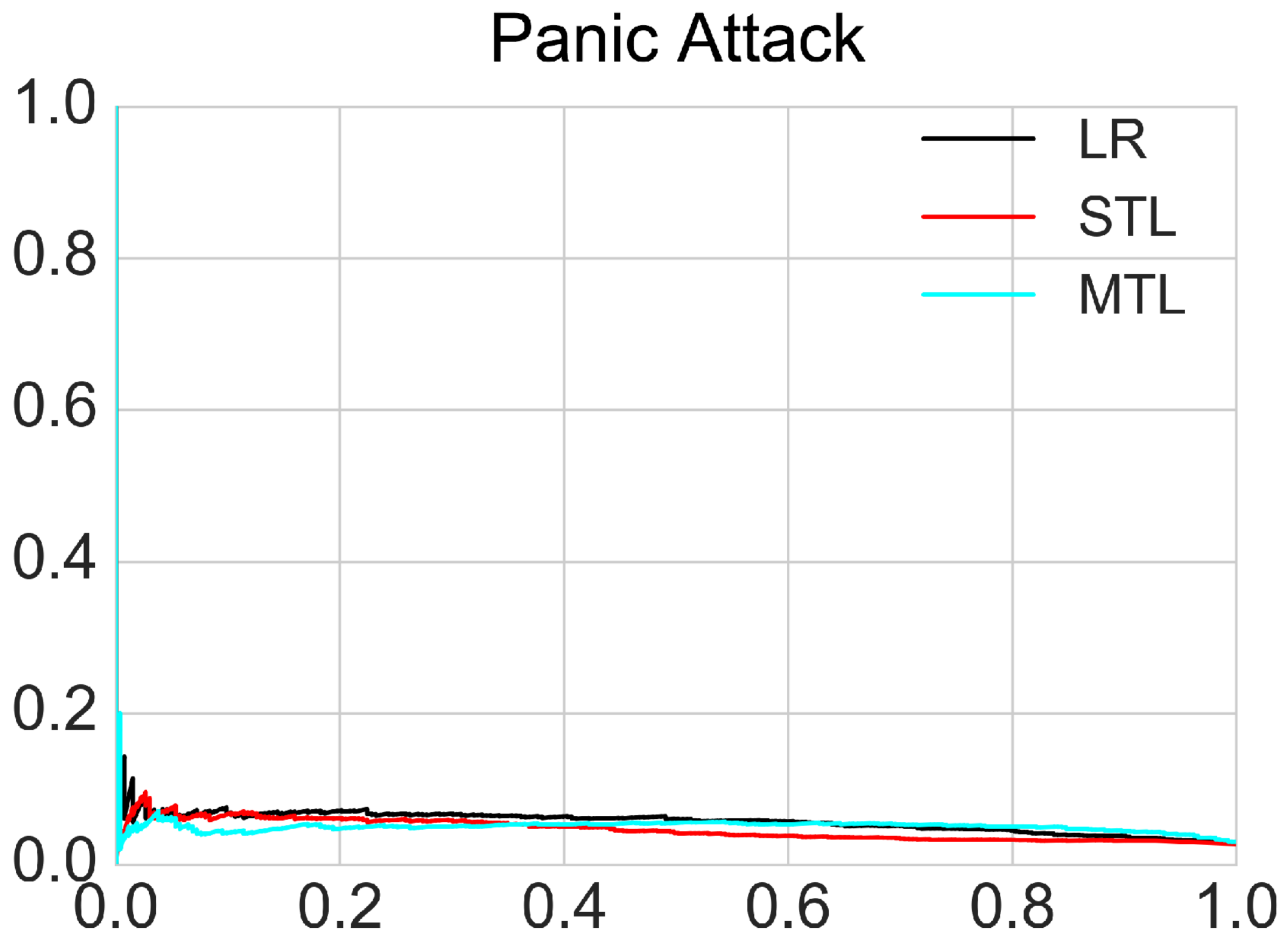}\\
        
\vspace{1em}
\includegraphics[width=0.3\textwidth,trim={0 12.25cm 0.1cm 0},clip]{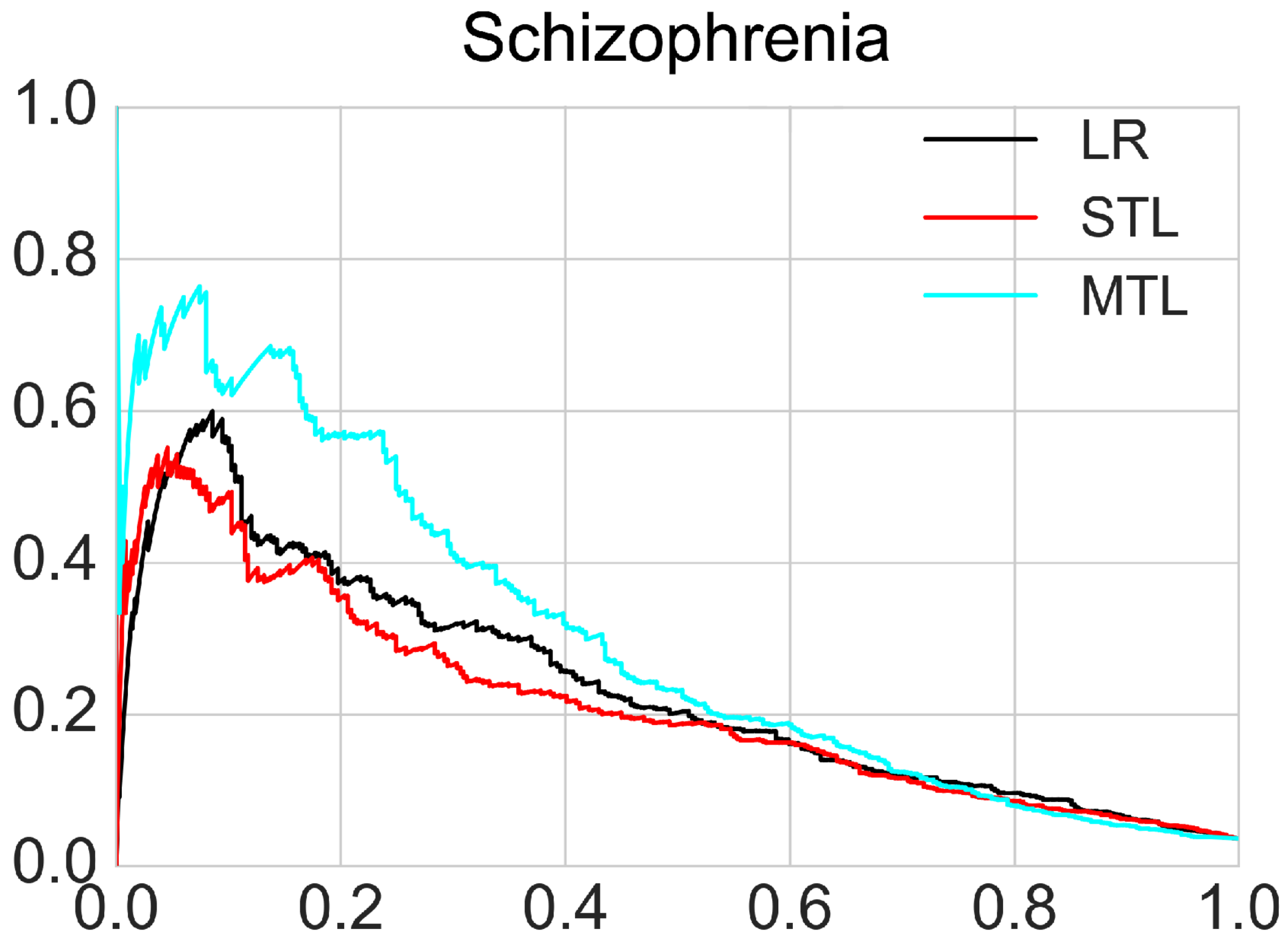} \hspace{1em}
		\includegraphics[width=0.3\textwidth,trim={0 12.25cm 0.1cm 0},clip]{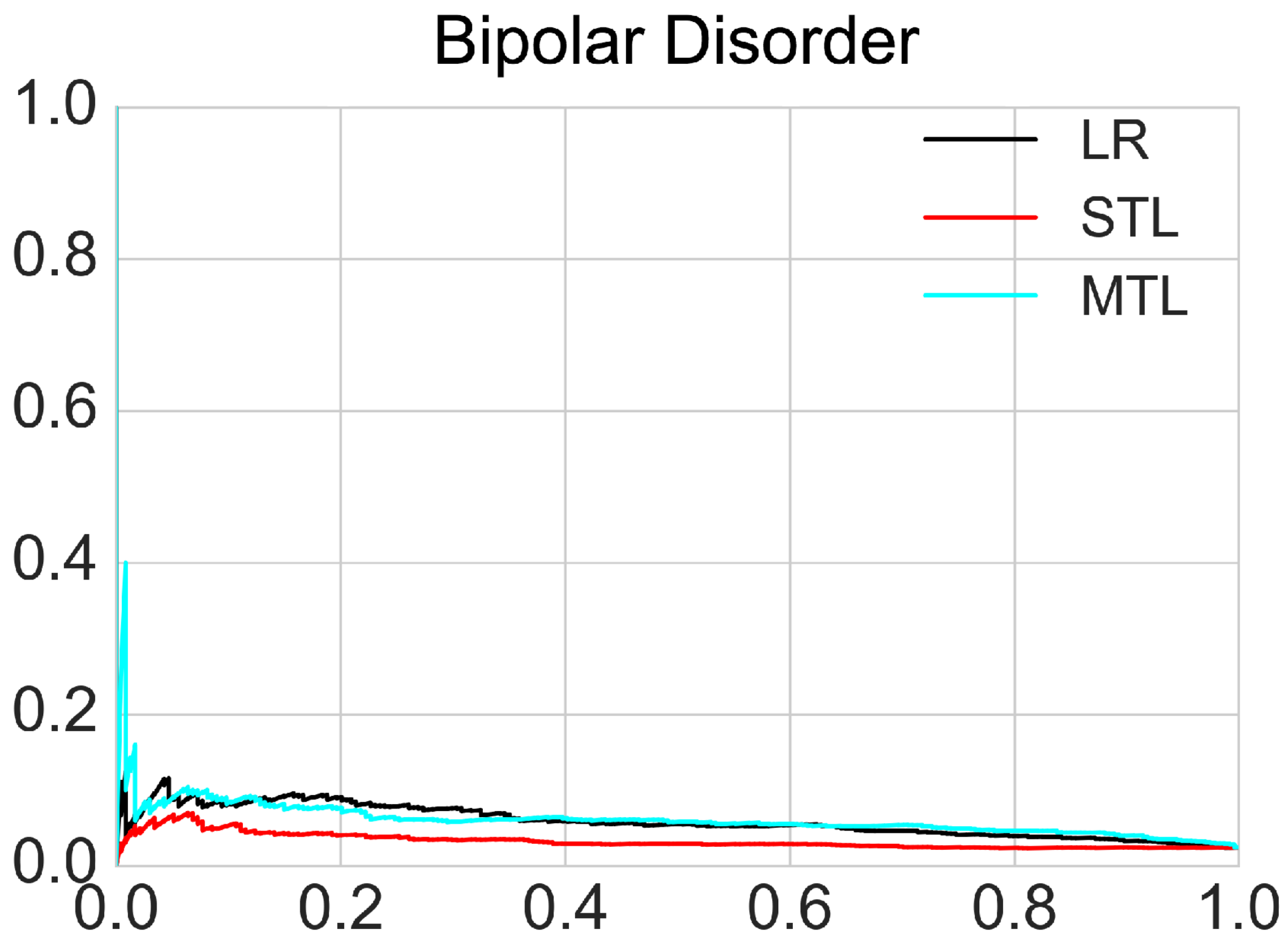}
      \hspace{1em}  \includegraphics[width=0.3\textwidth,trim={0 12.25cm 0.1cm 0},clip]{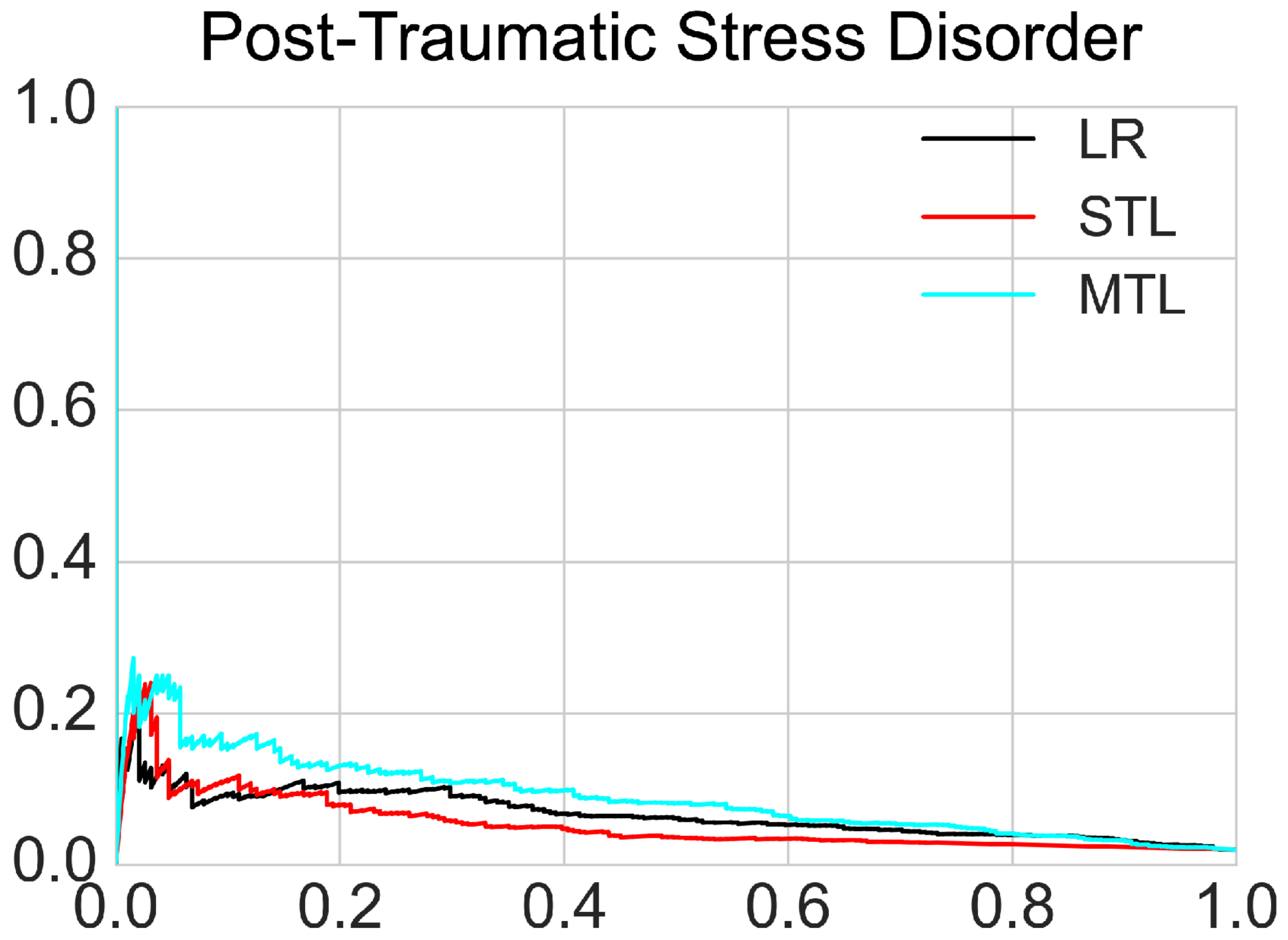}
		\captionof{figure}{Precision-recall curves for predicting each condition.
      \label{pic:precrec_curves}}
	\end{center}
\end{figure*}

Figure \ref{pic:auc} shows the AUC-score of each model for each task separately, and Figure \ref{pic:tpr} the true positive rate at a low false positive rate of 0.1.  Precision-recall curves for model/task are in Figure \ref{pic:precrec_curves}.  \STL{} is a multilayer perceptron with two hidden layers (with a similar number of parameters as the proposed \MTL{} model).  The MTL +gender and MTL models predict all tasks simultaneously, but are only evaluated on the main respective task.


Both AUC and TPR (at FPR=0.1) demonstrate that single-task models models do not perform nearly as well as multi-task models or logistic regression.  This is likely because the neural networks learned by \STL{} cannot be guided by the inductive bias provided by \MTL{} training.  Note, however, that \STL{} and \MTL{} are often times comparable in terms of F1-score, where false positives and false negatives are equally weighted.  

As shown Figure \ref{pic:auc}, multi-task suicide predictions reach an AUC of 0.848, and predictions for anxiety and schizophrenia are not far behind.  Interestingly however, schizophrenia stands out as being the only condition to be best predicted with a single-task model.  MTL models show improvements over STL and LR models for predicting suicide, neuroatypicality, depression, anxiety, panic, bipolar disorder, and PTSD.  The inclusion of gender in the MTL models leads to direct gains over an LR baseline in predicting anxiety disorders: anxiety, panic, and PTSD.

In Figure \ref{pic:tpr}, we illustrate the true positive rate -- that is, how many cases of mental health conditions that we correctly predict -- given a low false positive rate -- that is, a low rate of predicting people have mental health conditions when they do not.  This is particularly useful in clinical settings, where clinicians seek to minimize over-diagnosing.  In this setting, MTL leads to the best performance across the board, for all tasks under consideration:  Neuroatypicality, suicide, depression, anxiety, eating, panic, schizophrenia, bipolar disorder, and PTSD.  Including gender in MTL further improves performance for neuroatypicality, suicide, anxiety, schizophrenia, bipolar disorder, and PTSD.



\paragraph{MTL Leveraging Comorbid Conditions Improves Prediction Accuracy}

We find that the prediction of the conditions with the least amount of data -- \emph{bipolar disorder} and \emph{PTSD} -- are significantly improved by having the model also predict comorbid conditions with substantially more data:  \emph{depression} and \emph{anxiety}.  We are able to increase the AUC for predicting PTSD to 0.786 by \MTL, from 0.770 by \LR, whereas \STL{} fails to perform as well with an AUC of 0.667.  Similarly for predicting bipolar disorder (\MTL:0.723, \LR:0.752, \STL:0.552) and panic attack (\MTL:0.724, \LR:0.713, \STL:0.631).




These differences in AUC are significant at $p=0.05$ according to bootstrap sampling tests with 5000 samples.  The wide difference between MTL and STL can be explained in part by the increased feature set size -- MTL training may, in this case, provide a form of regularization that \STL{} cannot exploit.  Further, modeling the common mental health conditions with the most data (depression, anxiety) helps in pulling out more rare conditions comorbid with these common health conditions.  This provides evidence that an MTL model can help in predicting elusive conditions by using large data for common conditions, and a small amount of data for more rare conditions.


\paragraph{Utility of Authorship Attributes}
Figures \ref{pic:auc} and \ref{pic:tpr} both suggest that adding gender as an auxiliary task leads to more predictive models, even though the difference is not statistically significant for most tasks. This is consistent with the findings in previous work \cite{Volkova:ea:2013exploring,hovy2015demographic}.
Interestingly, though, the \MTL{} model is worse at predicting gender itself. While this could be a direct result of data sparsity (recall that we have only a small subset annotated for gender), which could be remedied by annotating additional users for gender, this appears unlikely given the other findings of our experiments, where \MTL{} helped in specifically these sparse scenarios. 

However, it has been pointed out by \newcite{caruana1996algorithms} that not all tasks benefit from a \MTL{} setting in the same way, and that some tasks serve purely auxiliary functions. Here, gender prediction does not benefit from including mental conditions, but helps vice versa.
In other words, predicting gender is qualitatively different from predicting mental health conditions: it seems likely that the signals for anxiety ares much more similar to the ones for depression than for, say, being male, and can therefore add to detecting depression. However, the distinction between certain conditions does not add information for the distinction of gender.
The effect may also be due to the fact that these data were constructed with inferred gender (used to match controls), so there might be a degree of noise in the data.

\paragraph{Choosing Tasks}

Although \MTL{} tends to dominate \STL{} in our experiments, it is not clear whether modeling several tasks provide a beneficial bias in \MTL{} models in general, or if there exists  specific subsets of auxiliary tasks that are most beneficial for predicting suicide risk and related mental health conditions.   We perform ablation experiments by training \MTL{} models on  a subset of auxiliary tasks, and prediction for a single main task.  We focus on four conditions to predict well: suicide attempt, anxiety, depression, and bipolar disorder.  For each main task, we vary the auxiliary tasks we train the \MTL{} model with.  Since considering all possible subsets of tasks is combinatorily unfeasible, we choose the following task subsets as auxiliary:

\begin{itemize}
\itemsep-0.5em
\item \emph{all}: all mental conditions along with gender
\item \emph{all conds}: all mental conditions, no gender
\item \emph{neuro}: only neurotypicality
\item \emph{neuro+mood}: neurotypicality, depression, and bipolar disorder (mood disorders)
\item \emph{neuro+anx}: neurotypicality, anxiety, and panic attack (anxiety conditions)
\item \emph{neuro+targets}: neurotypicality, anxiety, depression, suicide attempt, bipolar disorder
\item \emph{none}: no auxiliary tasks, equivalent to \STL{}
\end{itemize}

Table \ref{table:ablation} shows AUC for the four prediction tasks with different subsets of auxiliary tasks.  Statistically significant improvements over the respective LR baselines are denoted by superscript.  Restricting the auxiliary tasks to a small subset tends to hurt performance for most tasks, with exception to \emph{bipolar}, which benefits from the prediction of depression and suicide attempt. All main tasks achieve their best performance using the full set of additional tasks as auxiliary.  This suggests that the biases induced by predicting different kinds of mental conditions are mutually beneficial -- e.g., multi-task models that predict suicide attempt may also be good at predicting anxiety.

Based on these results, we find it useful to think of \MTL{}  
as a framework to leverage auxiliary tasks as regularization to effectively combat data paucity and less-than-trustworthy labels.  As we have demonstrated, this may be particularly useful when predicting mental health conditions and suicide risk.

\renewcommand{\arraystretch}{1.25}
\begin{table}[t]
\small
\begin{center}
\begin{tabular}{@{}r||l|l|l|l|@{}}
\multirow{5}{5em}{\bf Auxiliary Tasks} & \multicolumn{4}{c}{\bf{Main Task}} \\\cline{2-5}
 & \multicolumn{4}{c}{\vspace{-.5em}} \\
& \rot{\sc anxiety} & \rot{\multirow{2}{*}{\sc bipolar}} & \rot{\multirow{2}{*}{\sc depression}} & \rot{\multirow{2}{3em}{\sc suicide attempt}}\\
 & \multicolumn{4}{c}{\vspace{-.5em}} \\\hline
 

\emph{all} & $0.813^{*\dagger}$ & $0.752^{*\dagger}$ & $0.769^{\dagger}$ & $0.835^{*\dagger}$ \\
\emph{all conds} & 0.786 & $0.743^{\dagger}$ & $0.772^{\dagger}$ & $0.833^{*\dagger}$ \\
\emph{neuro} & 0.763 & $0.740^{\dagger}$ & 0.759 & 0.797 \\
\emph{neuro+mood} & 0.756 & $0.742^{\dagger}$ & 0.761 & 0.804 \\
\emph{neuro+anx} & 0.770 & $0.744^{\dagger}$ & 0.746 & 0.792 \\
\emph{neuro+targets} & 0.750 & $0.747^{\dagger}$ & 0.764 & 0.817 \\
\hline 
\emph{none (\STL)} & 0.777 & 0.552 & 0.749 & 0.810 \\
\emph{\LR} & 0.791 & $0.723^{\dagger}$ & 0.763 & 0.817 \\
\end{tabular}
\end{center}
\caption{Test AUC when predicting \emph{Main Task} after training to predict a subset of auxiliary tasks.  Significant improvement over \LR{} baseline at $p=0.05$ is denoted by $^*$, and over no auxiliary tasks (\STL) by $^\dagger$.}
\label{table:ablation}
\end{table}

\section{Discussion: Multi-task Learning}

Our results indicate that an \MTL{} framework can lead to significant gains over single-task models for predicting suicide risk and several mental health conditions. 
We find benefit from predicting related mental conditions and demographic attributes simultaneously.

We experimented with all the optimizers that Keras provides, and found that Adagrad seems to converge fastest to a good optimum, although all the adaptive learning rate optimizers (such as Adam, etc.) tend to converge quickly.  This indicates that the gradient is significantly steeper along certain parameters than others.  Default stochastic gradient descent (SGD) was not able to converge as quickly, since it is not able to adaptively scale the learning rate for each parameter in the model -- taking too small steps in directions where the gradient is shallow, and too large steps where the gradient is steep.
We further note an interesting behavior: all of the adaptive learning rate optimizers yield a strange ``step-wise'' training loss learning curve, which hits a  plateau, but then  drops after about 900 iterations, only to hit another plateau, and so on.
Obviously, we would prefer to have a smooth training loss curve.  We can indeed achieve this using SGD, but it takes much longer to converge than, for example, Adagrad.  This suggests that a well-tuned SGD would be the best optimizer for this problem, a step that would require some more experimentation and is left for future work.  

We also found that feature counts have a pronounced effect on the loss curves: Relative feature frequencies yield models that are much easier to train than raw feature counts.

Feature representations are therefore another area of optimization, e.g., different ranges of character $n$-grams (e.g., $n > 5$) and unigrams.  We used character 1-to-5-grams, since we believe that these features generalize better to a new domain (e.g., Facebook) than word unigrams.  However, there is no fundamental reason not to choose longer character $n$-grams, other than time constraints in regenerating the data, and  accounting for overfitting with proper regularization.

Initialization is a decisive factor in neural models, and \newcite{goldberg2015primer} recommends repeated restarts with differing initializations to find the optimal model. 
In an earlier experiment, we tried initializing a MTL model (without task-specific hidden layers) with pretrained word2vec embeddings of unigrams trained on the Google News $n$-gram corpus. However, we did not notice an improvement in F-score. This could be due to the other factors, though, such as feature sparsity.

Table \ref{table:sweep} shows parameters sweeps with hidden layer width 256, training the \MTL{} model on the social media data with character trigrams as input features.  The sweet spots in this table may be good starting points for training models in future work.

\begin{table}
\small
\begin{center}
\begin{tabular}{ r|c|r|c|r|c }
Learning & Loss & L2 & Loss & Hidden & Loss \\
Rate & & & & Width & \\
 \hline
 $10^{-4}$ & 5.1 & $10^{-3}$ & 2.8 & 32 & 3.0 \\
 $5*10^{-4}$ & 2.9 & $5*10^{-3}$ & 2.8 & 64 & 3.0 \\
 $10^{-3}$ & 2.9 & $10^{-2}$ & 2.9 & 128 & 2.9 \\
 $5*10^{-3}$ & 2.4 & $5*10^{-2}$ & 3.1 & 256 & 2.9 \\
 $10^{-2}$ & 2.3 & 0.1 & 3.4 & 512 & 3.0 \\
 $5*10^{-2}$ & 2.2 & 0.5 & 4.6 & 1024 & 3.0\\
 0.1 & 20.2 & 1.0 & 4.9 & & \\
 \hline
\end{tabular}
\end{center}
\caption{Average dev loss over epochs 990-1000 of joint training on all tasks as a function of different learning parameters.  Optimized using Adagrad with hidden layer width 256.}
\label{table:sweep}
\end{table}


\section{Related Work}
MTL was introduced by \newcite{caruana1993multitask}, based on the observation that humans rarely learn things in isolation, and that it is the similarity between related tasks that helps us get better.

Some of the first works on MTL were motivated by medical risk prediction \cite{caruana1996using}, and it is now being rediscovered for this purpose \cite{lipton2016learning}. The latter use a long short-term memory (LSTM) structure to provide several medical diagnoses from health care features (yet no textual or demographic information), and find small, but probably not significant improvements over a structure similar to the \STL{} we use here.

The target in previous work was medical conditions as detected in patient records, not mental health conditions in social text.  The focus in this work has been on the possibility of predicting suicide attempt and other mental health conditions using social media text that a patient may already be writing, without requiring full diagnoses.

The framework proposed by~\newcite{collobert:ea:2011} allows for predicting any number of NLP tasks from a convolutional neural network (CNN) representation of the input text.  The model we present is much simpler: A feed-forward network with $n$-gram input layer, and we demonstrate how to constrain $n$-gram embeddings for clinical application.  Comparing with further models is possible, but distracts from the question of whether MTL training can help in this domain.  As we have shown, it can.

\section{Conclusion and Future Work}
In this paper, we develop neural \MTL{} models for 10 prediction tasks (suicide, seven mental health conditions, neurotypicality, and gender). We compare their performance with \STL{} models trained to predict each task independently.

Our results show that an \MTL{} model with all task predictions performs significantly better than other models, reaching $0.846$ TPR for neuroatypicality where FPR=$0.1$, and AUC of $0.848$, TPR of $0.559$ for suicide.  Due to the nature of \MTL, we find additional contributions that were not the original goal of this work: Pronounced gains in detecting anxiety, PTSD, and bipolar disorder.  MTL  predictions for anxiety, for example, reduce the error rate from a single-task model by up to $11.9$\%.

We also investigate the influence of model depth, comparing to progressively deeper \STL{} feed-forward networks with the same number of parameters. We find: (1) Most of the modeling power stems from the expressivity conveyed by deep architectures. (2) Choosing the right set of auxiliary tasks for a given mental condition can yield a significantly better model.
(3) The \MTL{} model dramatically improves for conditions with the smallest amount of data.
(4) Gender prediction does not follow the two previous points, but improves performance as an auxiliary task.

Accuracy of the \MTL{} approach is not yet ready to be used in isolation in the clinical setting.  However, our experiments suggest this is a promising direction moving forward.  There are strong gains to be made in using multi-task learning to aid clinicians in their evaluations, and with further partnerships between the clinical and machine learning community, we foresee improved suicide prevention efforts.
\NoteMM{the world will be saved, etc.}

\section*{Acknowledgements}
Thanks to Kristy Hollingshead Seitz, Glen Coppersmith, and H. Andrew Schwartz, as well as the organizers of the Johns Hopkins Jelinek Summer School 2016. We are also grateful for the invaluable feedback on MTL from Yoav Goldberg, Stephan Gouws, Ed Greffenstette, Karl Moritz Hermann, and Anders S\o{}gaard.  The work reported here was started at JSALT 2016, and was supported by JHU via grants from DARPA (LORELEI), Microsoft, Amazon, Google and Facebook.

\bibliographystyle{eacl2017}
\bibliography{eacl2017_multi}

\end{document}